\documentclass{article}

\usepackage{microtype}
\usepackage{graphicx}
\usepackage{subfigure}
\usepackage{booktabs} 
\usepackage{multirow}
\usepackage{makecell}

\usepackage[hyphens]{url}
\usepackage{hyperref}
\usepackage[hyphenbreaks]{breakurl}

\usepackage[accepted]{icml2023}

\usepackage{amsmath}
\usepackage{amssymb}
\usepackage{mathtools}
\usepackage{amsthm}

\usepackage[capitalize,noabbrev]{cleveref}

\theoremstyle{plain}

\theoremstyle{definition}

\theoremstyle{remark}

\usepackage[textsize=tiny]{todonotes}

\icmltitlerunning{Topological Feature Selection}

\begin{document}

\twocolumn[
\icmltitle{Topological Feature Selection}



\icmlsetsymbol{equal}{*}

\begin{icmlauthorlist}
\icmlauthor{Antonio Briola}{a1,a2}
\icmlauthor{Tomaso Aste}{a1,a2,a3}
\end{icmlauthorlist}

\icmlaffiliation{a1}{Department of Computer Science, University College London, London, United Kingdom}
\icmlaffiliation{a2}{Centre for Blockchain Technologies, University College London, London, United Kingdom}
\icmlaffiliation{a3}{Systemic Risk Centre, London School of Economics, London, United Kingdom}

\icmlcorrespondingauthor{Antonio Briola}{antonio.briola.20@ucl.ac.uk}

\icmlkeywords{Topology, Feature Selection, Network Science, Complex Systems, ICML}
\vskip 0.3in
]



\printAffiliationsAndNotice{\icmlEqualContribution} 

\begin{abstract}
In this paper, we introduce a novel unsupervised, graph-based filter feature selection technique which exploits the power of topologically constrained network representations. We model dependency structures among features using a family of chordal graphs (i.e. the Triangulated Maximally Filtered Graph), and we maximise the likelihood of features' relevance by studying their relative position inside the network. Such an approach presents three aspects that are particularly satisfactory compared to its alternatives: (i) it is highly tunable and easily adaptable to the nature of input data; (ii) it is fully explainable, maintaining, at the same time, a remarkable level of simplicity; (iii) it is computationally cheap. We test our algorithm on $16$ benchmark datasets from different application domains showing that it  outperforms or matches the current state-of-the-art under heterogeneous evaluation conditions. The code and the data to reproduce all the results presented in the current research work are available at \url{https://github.com/FinancialComputingUCL/Topological_Feature_Selection}.
\end{abstract}

\section{Introduction} \label{sec:Introduction}
In the era of big data, effective management of extensive feature spaces represents a genuine hurdle for scientists and practitioners. Only some features have significant relevance in real-world data, while the redundancy of the remaining ones actively inflates data-driven models' complexity. The search for always new and increasingly performing dimensionality reduction algorithms is hence justified by four primary needs: (i) reduce data maintenance costs \cite{wu2016building, eichelberger2017experiences}; (ii) reduce the impact of the `curse of dimensionality' \cite{donoho2000high, poggio2017and} on data-driven models; (iii) increase data-driven models' interpretability \cite{miao2016survey}; (iv) reduce energy costs for models' training \cite{strubell2019energy, patterson2021carbon}.

Dimensionality reduction addresses all these challenges by decreasing the complexity of the feature space while minimizing the information loss \cite{miner2012practical}. Such a field can be further specified into two macro areas: (i) feature extraction \cite{mingqiang2008survey} and (ii) feature selection \cite{chandrashekar2014survey}. Feature extraction techniques originate new features by transforming the original ones into a space with a different dimensionality and then choosing linear or non-linear combinations of them. On the other hand, feature selection techniques directly choose a subset of features from the original set while maintaining their physical meaning. Usually, they represent a ‘block’ of a more complex pipeline including also a classification (or regression) step. The level of contamination among ‘blocks’ is more or less evident depending on the nature of the feature selection algorithm. Indeed, feature selection techniques are classified as (i) supervised \cite{huang2015supervised}; (ii) unsupervised \cite{solorio2020review}: and (iii) semi-supervised \cite{sheikhpour2017survey}. In the supervised case, features' relevance is usually assessed via their correlation degree with class labels or regression targets. These models take advantage of the learning performance of the classification (or regression) algorithm to refine the choice of the meaningful subset of features while maintaining, at the same time, ‘block’ independence from it. This loop interaction is convenient only in scenarios where retrieving labels and executing classification (or regression) tasks is computationally efficient. On the contrary, unsupervised feature selection is applied in scenarios where retrieving labels is costly. These algorithms select relevant features based on specific data properties (e.g. variance maximisation). Different classes of unsupervised feature selection approaches exist: (i) filters; (ii) wrappers; (iii) embedded methods. In filter-based methods, the feature selection stage is entirely independent from the classification (or regression) algorithm; in wrapper-based methods, the feature selection process takes advantage of classification (or regression) stage to improve its performance; in embedded methods, the feature selection step is embedded into the classification (or regression) algorithm in a way that the two ‘blocks’ take advantage from each other. Lastly, semi-supervised feature selection represents a hybrid approach with the highest potential applicability in real-world problems. Indeed, labels are often only partially provided, and semi-supervised learning techniques are specifically designed to learn from a reduced number of labelled data, efficiently handling, at the same time, a large number of unlabeled samples.

In its general formulation, given a starting set of features $F = \{f_1, \dots, f_n\}$, where $n \gg 1$ is the cardinality, dimensionality reduction is an \textit{NP-hard} problem \cite{guyon2008feature, roffo2015infinite} where the goal is selecting the optimal subset of features with cardinality $m \ll n$, among the ${n \choose m}$ possible combinations. Due to the exponentially increased time required to find the globally optimal solution, existing feature selection algorithms employ heuristic rules to find dataset-dependent sub-optimal solutions \cite{ge2016mctwo}. In this paper, we introduce ‘Topological Feature Selection’ (TFS), a novel unsupervised, graph-based filter algorithm for feature selection which (i) builds a topologically constrained network representation of raw features' dependency structure and (ii) exploits their relative position into the graph to reduce the input's dimensionality while maximising the likelihood of features’ relevance and minimising the information loss.

To prove the effectiveness of the proposed methodology, we test it against what is currently considered the state-of-the-art counterpart in unsupervised, graph-based filter feature selection approaches, i.e. Infinite Feature Selection ($\textrm{Inf-FS}_U$) \cite{roffo2020infinite}. The two feature selection algorithms are tested on $16$ benchmark datasets from different application domains. The proposed training/test pipeline and the statistical validation stage are designed to handle datasets' imbalance and evaluate results based on fair performance metrics. The results are clear-cut. In most cases, TFS outperforms or equalises its alternative, redefining or teeing state-of-the-art performances. Our contribution to the existing literature is relevant since we propose an extraordinarily flexible, computationally cheap and remarkably intuitive (compared to its alternative) unsupervised, graph-based filter algorithm for feature selection, guaranteeing complete control of the dimensionality reduction process by considering only the relative position of features inside well-known graph structures.

The rest of the paper is organised as follows. In Section \ref{sec:Data}, we describe the datasets used to prove the effectiveness of the TFS algorithm. In Section \ref{sec:Information_Filtering_Networks}, we review the theoretical foundations of TFS. In Section \ref{sec:Topologically_Constrained_Feature_Selection}, we describe in a detailed manner the TFS methodology. In Section \ref{sec:Experiments}, we describe the experimental protocols, while obtained results are presented in Section \ref{sec:Results}. Finally, in Section \ref{sec:Conclusions}, we discuss the meaning of our results and future research lines in this area.

\section{Methods}\label{sec:Data_and_Methods}

\subsection{Data} \label{sec:Data}
We prove TFS' effectiveness by running a battery of tests on $16$ benchmark datasets (all in a tabular format) belonging to $7$ different application domains (i.e. text data, face images data, hand written images data, biological data, digits data, spoken letters data and artificial data) (see Appendix \ref{Appendix_1} for further details). The data pre-processing pipeline consists of $3$ different steps: (i) data reading and format unification; (ii) training/validation/test splitting; (iii) constant features pruning. The first step allows to read data coming from different sources and unify the format. The second step organises each dataset into a training, validation and a test set. In the third step, non-informative, constant covariates are detected on the training set and permanently removed from the training, validation and the test set. In Appendix \ref{Appendix_2}, we report datasets’ specifics after the pre-processing step. We remark that most of the considered datasets are not affected by the constant features filtering step. Datasets are roughly balanced and the raw labels' distribution is  preserved through the execution of stratification \cite{zeng2000distribution} during the training/validation/test splitting stage.

\subsection{Information Filtering Networks} \label{sec:Information_Filtering_Networks}
Information Filtering Networks (IFNs) \cite{mantegna1999hierarchical, aste2005complex, tumminello2005tool, barfuss2016parsimonious, massara2017network} are an effective tool to represent and model dependency structures among variables characterising complex systems. In the past two decades, they have been extensively used in finance \cite{procacci2022portfolio, briola2022anatomy, wang2022dynamic, seabrook2022quantifying, wang2022sparsification, vidal2023ftx}, psychology \cite{christensen2018network, christensen2018networktoolbox}, medicine \cite{hutter2019multi, danoff2021genetic} and biology \cite{song2008correlation, song2012hierarchical}. Sometimes IFNs are also referred to as Correlation Networks (CNs). Such an association is, however, inaccurate. Indeed, the two methodologies slightly differ, with CNs being normally obtained by imposing a threshold that retains only the largest correlations among variables of the system, while IFNs being constructed imposing additional topological constraints (e.g. being a tree or a planar graph) and optimising specific global properties (e.g. likelihood) \cite{aste2022topological}. Both methodologies end in the determination of a sparse adjacency matrix, \textbf{\textit{A}}, representing relations among variables in the system with the fundamental difference that the former approach generates a disconnected graph, while the latter guarantees the connectedness. Based on the nature of the relationships to be modelled (i.e. linear, non-linear), one can choose different metrics to build the adjacency matrix \textbf{\textit{A}}. In most cases, \textbf{\textit{A}} is built on an arbitrary similarity matrix $\hat{\textbf{C}}$, which often corresponds to a correlation matrix. From a network science perspective, $\hat{\textbf{C}}$ can be considered as a fully connected graph where each variable of the system is represented as a node, and each pair of variables is joined by a weighted and undirected edge representing their similarity. Historically, the main IFNs were the Minimum Spanning Tree (MST) \cite{papadimitriou1998combinatorial, mantegna1999hierarchical} and the Planar Maximally Filtered Graph (PMFG) \cite{tumminello2005tool}. MSTs are a class of networks connecting all the vertices without forming cycles (i.e. closed paths of at least three nodes) while retaining the network's representation as simple as possible (i.e. representing only relevant relations among variables characterising the system under analysis) \cite{briola2022dependency}. Prim's algorithm for MST construction sorts all edges' weights (i.e. similarities) in descending order and adds the largest possible edge weight among two nodes in an iterative way. The resulting network has $n-1$ edges and retains only the most significant connections, assuring, at the same time, that connectedness' property is fulfilled \cite{christensen2018network}. Despite being a powerful method to capture meaningful relationships in network structures describing complex systems, MST presents some aspects that can be unsatisfactory. Paradoxically, the main limit is represented by its tree structure (i.e. it cannot contain cycles) which does not allow to represent direct relationships among more than two variables showing strong similarity. The introduction of the Planar Maximally Filtered Graph (PMFG) \cite{tumminello2005tool} overcomes such a shortcoming. Similarly to MST, also the PMFG algorithm sorts edge weights in descending order, incrementally adding the largest ones while imposing planarity \cite{christensen2018network}. A graph is planar if it can be embedded in a sphere without edges crossing. Thanks to this, the same powerful filtering properties of the MST are maintained, and, at the same time, extra links, cycles and cliques (i.e. complete subgraphs) are added in a controlled manner. The resulting network has $3n-6$ edges and is composed of three- and four-nodes cliques. A nested hierarchy emerges from these cliques \cite{song2011nested}: dimensionality is reduced in a deterministic manner while local information and the global hierarchical structure of the original network are retained. The PMFG represents a substantial step forward compared to the MST. However, it still presents two limits: (i) it is computationally costly and (ii) it is a non-chordal graph. A graph is said to be chordal if all cycles made of four or more vertices have a chord, reducing the cycle to a set of triangles. A chord is defined as an edge that is not part of the cycle but connects two vertices of the cycle itself. The advantage of chordal graphs is that they fulfill the independence assumptions of Markov (i.e., bidirectional or undirected relations) and Bayesian (i.e., directional relations) networks \cite{koller2009probabilistic, christensen2018network}. The Triangulated Maximally Filtered Graph (TMFG) \cite{massara2017network} has been explicitly designed to be a chordal graph while retaining the strengths of PMFG. The building process of TMFG (see Appendix \ref{Appendix_3} for further details) is based on a simple topological move that preserves planarity: it adds one node to the centre of three-nodes cliques by using a score function that maximises the sum of the weights of the three edges connecting the existing vertices. This addition transforms three-nodes cliques (i.e. triangles) into four-nodes cliques (i.e. tetrahedrons) characterised by a chord that forms two triangles and generates a chordal network \cite{christensen2018network}. Also in this case, the resulting network has $3n-6$ edges and is composed of three- and four-nodes cliques. TMFG has two main advantages compared to PMFG: (i) it can be used to generate sparse probabilistic models as a form of topological regularization \cite{aste2022topological} and (ii) it is computationally more efficient. On the other hand, the two main limitations of chordal networks are that (i) they may add unnecessary edges to satisfy the property of chordality and (ii) their building cost can vary based on the chosen optimization function.

\subsection{Topological Feature Selection} 
\label{sec:Topologically_Constrained_Feature_Selection}
Topological Feature Selection (TFS) algorithm is a graph-based filter method to perform feature selection in an unsupervised manner. Given a set of features $F = \{f_1, \dots, f_n\}$, where $n \gg 1$ is the cardinality, we build the adjacency matrix \textbf{\textit{A}} of the corresponding TMFG based on one of the following three metrics: (i) the Pearson's estimator of the correlation coefficient, (ii) the Spearman's rank correlation coefficient and (iii) the Energy coefficient (i.e. the weighted combination of two pairwise measures described later in this Section). Depending on the metric's formulation, it is possible to capture different kinds of interactions among covariates (e.g. linear or non-linear interactions).

The Pearson’s estimator of the correlation coefficient for the two covariates $f_i$ and $f_j$, is defined as:

\begin{equation}
    r_{f_i, f_j} = \frac{\sum_{s=1}^S ({f_{i,s}} - \hat{\mu}_{f_i}) ({f_{j,s}} - \hat{\mu}_{f_j})}{\hat{\sigma}_{f_i}\hat{\sigma}_{f_j}}
\end{equation}

where $S$ is the sample size, $f_{i,s}$ and $f_{j, s}$ are two sample points indexed with $s$, $\hat{\mu}$ is the sample mean and $\hat{\sigma}$ is the sample standard deviation. By definition, $r_{f_i, f_j}$ has values between $-1$ (meaning that the two features are completely, linearly anti-correlated), and $+1$ (meaning that the two features are completely, linearly correlated). When $r_{f_i, f_j} = 0$, the two covariates are said to be uncorrelated. The Person’s estimator of the correlation coefficient heavily depends on the distribution of the underlying data and may be influenced by outliers. In addition to this, it only captures linear dependency among variables, restricting its applicability to real-world problems where non-linear interactions are often relevant \cite{shirokikh2013computational}. 

The Spearman's rank correlation coefficient is based on the concept of `variables ranking'. Ranking a variable means mapping its realizations to an integer number that describes their positions in an ordered set. Considering a variable with cardinality $|s|$, this means assigning 1 to the realization with the highest value and $|s|$ to the realization with the lowest value.

The Spearman's rank correlation coefficient for the two covariates $f_i$ and $f_j$, is defined as the Pearson's correlation between the ranks of the variables:

\begin{equation}
    r_{s_{f_i, f_j}} = \frac{\sum_{s=1}^S (R_{f_{i, s}} - \hat{\mu}_{R_{f_i}})(R_{f_{j, s}} - \hat{\mu}_{R_{f_j}})}{\hat{\sigma}_{R_{f_i}}\hat{\sigma}_{R_{f_j}}}
\end{equation}

where $S$ is the sample size, $R_{f_{i, s}}$ and $R_{f_{j, s}}$ are the ranks of the two sample points indexed with $s$, $\hat{\mu}$ is the sample mean for $R_{f_i}$ and $R_{f_j}$ and $\hat{\sigma}$ is the sample standard deviation for $R_{f_i}$ and $R_{f_j}$. If there are no repeated data samples, a perfect Spearman's rank correlation (i.e. $r_{s_{f_i, f_j}} = 1$ or $r_{s_{f_i, f_j}} = -1$ occurs when each of the features is a perfect monotone function of the other). Spearman's rank correlation technique is distribution-free and allows to capture monotonic, but not necessarily linear, relationships among variables \cite{shirokikh2013computational}. 

The Energy coefficient is a metric introduced by \cite{roffo2015infinite}, and it is used as the primary benchmark for comparison between our method and the current state-of-the-art. It is a weighted combination of two different pairwise measures defined as follows:

\begin{equation} \label{eq:Energy_Coefficient}
    \phi_{f_i, f_j} = \alpha E_{f_i,f_j} + (1-\alpha) \rho_{f_i,f_j}
\end{equation}

where $E_{f_i,f_j} = \max(\hat{\sigma}_{f_i}, \hat{\sigma}_{f_j})$, with $\hat{\sigma}$ being the sample standard deviation computed on features $f_i$ and $f_j$ normalized to the range $[0,1]$, $\rho_{f_i,f_j} = 1 - |r_{s_{f_i, f_j}}|$ and $\alpha$ is a threshold value with a value $\in [0, 1]$. $\phi_{f_i, f_j} \in [0, 1]$ analyses two features distributions (i.e. $f_i$ and $f_j$) taking into account both their maximal dispersion (i.e. standard deviation) and their level of uncorrelation. Computing $\phi_{f_i, f_j}$ for all the features in $F$ in a pairwise manner, one can define a matrix which is $n \times n$ symmetric and completely characterized by $n(n-1)/2$ coefficients. For simplicity, we refer to this matrix as $\hat{\textbf{C}}$ too.

Once one of the above mentioned metrics is chosen and $\hat{\textbf{C}}$ is computed, TFS applies the standard TMFG algorithm defined in Appendix \ref{Appendix_3} on the corresponding fully connected graph, creating a sparse chordal network which is able (i) to retain useful relationships among features, (ii) prune the weakest ones, and (iii) express a significant level of information flow among input variables.

The last step toward the selection of the most relevant features, is represented by the choice of the right nodes inside the TMFG. In this sense, multiple approaches of increasing complexity can be formulated. In this paper, which is a foundational one, we study the relative position of the nodes in the network computing their degree centrality. Degree centrality is the simplest and least computationally intensive measure of centrality. Typically, all the other centrality measures are strictly related \cite{lee2006correlations, valente2008correlated}. Given the sparse adjacency matrix \textbf{\textit{A}} representing the TMFG, degree centrality of a node $v$ is denoted as $\deg(v)$ and represents the number of neighbours (i.e. how many edges a node has) of $v$ as follows:

\begin{equation}
    \deg(v) = \sum_{w=1}^n A_{f_v,f_w}
\end{equation}

where $n$ is the cardinality of $F$ and $f_v$ and $f_w$ are two features $\in F$. Despite its simplicity, degree centrality can be very illuminating and can be considered a crude measure of whether a node is influential or not in the TMFG (i.e. variables mostly contributing to the system's information flow). Once obtained $\deg(v) \; \forall v \in \textrm{TMFG}$, we rank these values in a descending order and we take the top $k$ central nodes, where $k$ is the cardinality of the features' subset we want to consider. 

\subsection{Benchmark method: Infinite Feature Selection ($\textrm{Inf-FS}_U$)} 
To prove the effectiveness of the proposed methodology, we test the TFS algorithm against the current state-of-the-art counterpart in unsupervised, graph-based filter feature selection techniques, i.e. Infinite Feature Selection ($\textrm{Inf-FS}_U$) \cite{roffo2020infinite}.~\footnote{The Python implementation of $\textrm{Inf-FS}_U$ algorithm used in this paper can be reached at \url{https://github.com/fullyz/Infinite-Feature-Selection}.} $\textrm{Inf-FS}_U$ represents features as nodes of a graph and relationships among them as weighted edges. Weights are computed as per in Equation \ref{eq:Energy_Coefficient}. Each path of a given length over the network is seen as a potential set of relevant features. Therefore, varying paths and letting them tend to an infinite number permits the investigation of the importance of each possible subset of features. Based on this, assigning a score to each feature and ranking them in descendant order allows us to perform feature selection effectively. It is worth noting that $\textrm{Inf-FS}_U$ has a computational complexity equal to $\approx \mathcal{O}(n^3)$. In contrast, TFS has a computational complexity equal to $\approx \mathcal{O}(n^2)$, with $n$ being the number of the features.

\subsection{Experiments} \label{sec:Experiments}
Once defined the training, validation and test set for each benchmark dataset, model hyper-parameters (see Table \ref{tab:Hyperparameters_Table}) are optimised adopting a parallel grid search approach. For $\textrm{Inf-FS}_U$, the first hyper-parameter to be optimised is $\alpha$, which can take values between $0$ and $1$. To tune this parameter, we use a range of equally spaced realisations between $0.1$ and $1.0$, all at a distance of $0.1$. The second hyper-parameter to be optimised is $\theta$ and represents a regularisation factor which, in the original paper \cite{roffo2015infinite}, has a fixed value equal to $0.9$. Here, instead, we tune this parameter in the same way as $\alpha$. In the case of TFS, the first hyper-parameter to be optimised is the metric used in the building process of the initial fully connected graph. As reported in Section \ref{sec:Topologically_Constrained_Feature_Selection}, we test three different metrics: (i) the Pearson's estimator of the correlation coefficient, (ii) the Spearman's rank correlation coefficient and  (iii) the Energy coefficient. The second hyper-parameter to be tuned is a boolean value which regulates the chance to use the coefficients mentioned above in a squared form. It is worth mentioning that, if the Energy metric is chosen, the corresponding $\hat{\textbf{C}}$ is never squared since it already contains only positive values. The last hyper-parameter to be optimised is $\alpha$, and it should be considered only when the Energy coefficient is chosen as a metric. The meaning of this hyper-parameter is the same as its homologous in the $\textrm{Inf-FS}_U$ model. All the models are finally evaluated on feature subsets with cardinalities $\in [10, 50, 100, 150, 200]$.

\begin{table}[h]
\centering
\caption{Model dependent hyper-parameters search spaces. In the case of the TFS algorithm, the $\dagger$ symbol indicates that, if the Energy metric is chosen, the corresponding $\hat{\textbf{C}}$ is never squared ($\hat{\textbf{C}}$ already contains only positive values). The $\ddagger$ symbol, on the other hand, indicates that $\alpha$ parameter should be considered only when the Energy coefficient is chosen as a metric.}
\vskip 0.1in
\label{tab:Hyperparameters_Table}
\scalebox{0.85}{
\begin{tabular}{@{}cc@{}}
\toprule
\textbf{Model} & \textbf{Hyper-parameters}                                      \\ \midrule
$\textrm{Inf-FS}_U$   & \begin{tabular}[c]{@{}c@{}}$\alpha$: {[}0.1: 0.1: 1.0{]}\\ $\theta$: {[}0.1: 0.1: 1.0{]}\end{tabular}                                              \\ \midrule
TFS & \begin{tabular}[c]{@{}c@{}}metric: {[}Pearson, Spearman, Energy{]}\\ square$^\dagger$: {[}True, False{]}\\ $\alpha^\ddagger$: {[}0.1: 0.1: 1.0{]}\end{tabular} \\ \bottomrule
\end{tabular}
}
\end{table}

For each hyper-parameter combination, a stratified $k$-fold cross-validation with $k=3$ is performed on the training set. The value of $k$ is chosen to take into account labels' distributions reported in Appendix \ref{Appendix_2}. Results' reproducibility and a fair comparison between models are guaranteed by fixing the random seed for each step of the training/validation/test pipeline. 

The meaningfulness of each subset of features chosen by the two algorithms is evaluated based on the classification performance achieved by three classification algorithms: (i) Linear Support Vector Classifier (LinearSVM); (ii) \textit{k}-Nearest Neighbors Classifier (KNN); (iii) Decision Tree Classifier (Decision Tree). LinearSVM is a sparse kernel-based method designed to convert non-linearly separable problems in the low-dimensional space, into linearly separable problems in the higher-dimensional space, thereby achieving classification \cite{han2022data}. KNN is a lazy learning algorithm, which classifies test instances evaluating their distance from the nearest \textit{k} training samples stored in an \textit{n}-dimensional space (where \textit{n} is the number of dataset's covariates) \cite{han2022data}. Finally, Decision Tree is a flowchart-like tree structure computed on training instances which classifies test samples tracing a path from the root to a leaf node holding the class prediction \cite{han2022data}. The inherently different nature of the three classifiers prevents from obtaining biased results for the two feature selection approaches. More details about chosen classifiers and their implementations are reported in Appendix \ref{Appendix_4}. Learning performances are evaluated based on three different metrics: (i) the Balanced Accuracy score (BA) \cite{mosley2013balanced, kelleher2020fundamentals}; (ii) the F1 score (F1); (iii) the Matthews Correlation Coefficient (MCC) \cite{baldi2000assessing, gorodkin2004comparing, jurman2012comparison}. We use the BA score for the hyper-parameters optimization process and as the reference metric to present results in Section \ref{sec:Results}. The BA score for the multi-class case is defined as:

\begin{equation} \label{eq:BA_multi}
    BA = \frac{1}{|Z|} \left(  \sum_{z \in Z} \frac{\textrm{TP}_z}{\textrm{TP}_z+\textrm{FN}_z} + \frac{\textrm{TN}_z}{\textrm{TN}_z+\textrm{FP}_z}   \right).
\end{equation}

$\textrm{TP}$ is the number of outcomes where the model correctly classifies a sample as belonging to a positive class, when in fact it does belong to that class. $\textrm{TN}$ is the number of outcomes where the model correctly classifies a sample as belonging to a negative class, when in fact it does not belong to that class. $\textrm{FP}$ is the number of outcomes where the model incorrectly classifies a sample as belonging to a positive class, when in fact it does not belong to that class. $\textrm{FN}$ is the number of outcomes where the model incorrectly classifies a sample as belonging to a negative class, when in fact it belongs to a positive class. $|Z|$ indicates the cardinality of the set of different classes.

General formulations for the F1 score, and the MCC are reported in Appendix \ref{Appendix_6} together with an extended version of results described later in this Section.

For each model and for each subset's cardinality, the hyper-parameters configuration which maximises the BA score while minimising the number of parameters is applied to test datasets. In order to assess the robustness of our findings, we test if the results achieved by the two feature selection approaches are statistically different. To achieve this goal we use an improved version of the classic $5\times2$ cv paired \textit{t}-test \cite{dietterich1998approximate}.  The test is constructed as follows. Given two classifiers $A$ and $B$ and a dataset $D$, $D$ is first randomly split into two balanced subsets $D_1$, $D_2$ (one for training and one for test). Both $A$ and $B$ are then estimated on $D_1$ and evaluated on $D_2$ obtaining performance measures $a_1, b_1$. The roles of the datasets are then switched by estimating $A$ and $B$ on $D_2$ and evaluating on $D_1$ which results in further performance measures $a_2, b_2$. The random division of $D$ is performed for a total of $5$ times, obtaining the matched performance evaluations $\{a_1, b_1\}, \{a_2, b_2\} \dots, \{a_{10}, b_{10}\}$. The test statistic \textit{t} is then computed as follows:

\begin{equation}
    t = \frac{\sqrt{10} \bar d}{\hat{\sigma}},
\end{equation}

where $d_h=a_h-b_h$ for $ h=1,\dots, 10$, is the difference between the matched performances metrics of the two classifiers, $\bar d = \frac{1}{10}\sum_{h=1}^{10} d_h$ and $\hat{\sigma}^2 = \frac{1}{10} \sum_{h=1}^{10}(d_h - \bar d)^2$. \textit{t} follows a \textit{t}-distribution with $9$ degrees of freedom and the null hypothesis is that \textit{the two classifiers $A$ and $B$ are not statistically different in their performances}. Starting from this basic formulation of the $5\times2$ cv paired \textit{t}-test, we simply increased the number of iterations, making it a $15\times2$ cv paired \textit{t}-test, in order to increase the statistical robustness of the achieved results. Also in this case, results reproducibility is guaranteed through a strict control of random seeds.

\section{Results} \label{sec:Results}

For both $\textrm{Inf-FS}_U$ and TFS, for each one of the three considered classifiers and for each feature subset cardinality $\in [10, 50, 100, 150, 200]$, results obtained running the hyper-parameter optimisation pipeline (see Section \ref{sec:Experiments}) are reported in Appendix \ref{Appendix_5}.

Table \ref{tab:local_bests_results} reports out-of-sample Balanced Accuracy scores obtained using subset's cardinality-dependent optimal hyper-parameter configurations for LinearSVM, KNN and Decision Tree classifier respectively. For each dataset, we highlight in bold the best achieved result. If one classifier performs equally across multiple subsets' cardinalities, the winning configuration is the one which minimises the subset's cardinality itself. If one classifier performs equally under the two feature selection schema, the winning feature selection approach is the one which minimises the computational complexity (i.e. TFS). 

To compare $\textrm{Inf-FS}_U$ and TFS, we consider three different measures: (i) the number of times a classifier achieves optimal results under each feature selection schema; (ii) the cross-datasets average balanced accuracy score; (iii) the cross-datasets average maximum drawdown ratio (i.e. the difference between the highest and the lowest achieved result).

\begin{table*}

\caption{Subset size-dependent, out-of-sample balanced accuracy scores using a LinearSVM, KNN and Decision Tree classifiers. For each dataset, we boldly highlight the combination between feature selection schema and classifier producing the best out-of-sample result. For each subset size, we report, in the last row, the number of times a feature selection approach outperforms the other across datasets.}
\label{tab:local_bests_results}
\vspace{0.1in}

\begin{minipage}[t]{0.4\linewidth}
\centering
\scalebox{0.52}{
\begin{tabular}{c|cccccccccc}
\hline
 &
  \multicolumn{10}{c}{\textbf{LinearSVM}} \\ \hline
 &
  \multicolumn{2}{c|}{\textbf{10}} &
  \multicolumn{2}{c|}{\textbf{50}} &
  \multicolumn{2}{c|}{\textbf{100}} &
  \multicolumn{2}{c|}{\textbf{150}} &
  \multicolumn{2}{c}{\textbf{200}} \\ \cline{2-11} 
 &
  \textbf{$\textrm{Inf-FS}_U$} &
  \multicolumn{1}{c|}{\textbf{TFS}} &
  \textbf{$\textrm{Inf-FS}_U$} &
  \multicolumn{1}{c|}{\textbf{TFS}} &
  \textbf{$\textrm{Inf-FS}_U$} &
  \multicolumn{1}{c|}{\textbf{TFS}} &
  \textbf{$\textrm{Inf-FS}_U$} &
  \multicolumn{1}{c|}{\textbf{TFS}} &
  \textbf{$\textrm{Inf-FS}_U$} &
  \textbf{TFS} \\ \hline
PCMAC &
  0.52 &
  \multicolumn{1}{c|}{0.50} &
  0.57 &
  \multicolumn{1}{c|}{0.67} &
  0.59 &
  \multicolumn{1}{c|}{0.70} &
  0.61 &
  \multicolumn{1}{c|}{\textbf{0.71}} &
  0.62 &
  0.69 \\
RELATHE &
  0.47 &
  \multicolumn{1}{c|}{0.49} &
  0.43 &
  \multicolumn{1}{c|}{\textbf{0.53}} &
  0.51 &
  \multicolumn{1}{c|}{0.53} &
  0.44 &
  \multicolumn{1}{c|}{0.49} &
  0.53 &
  0.53 \\
COIL20 &
  0.52 &
  \multicolumn{1}{c|}{0.63} &
  0.77 &
  \multicolumn{1}{c|}{0.90} &
  0.84 &
  \multicolumn{1}{c|}{0.92} &
  0.90 &
  \multicolumn{1}{c|}{0.94} &
  0.94 &
  \textbf{0.96} \\
ORL &
  0.40 &
  \multicolumn{1}{c|}{0.44} &
  0.63 &
  \multicolumn{1}{c|}{0.88} &
  0.72 &
  \multicolumn{1}{c|}{0.89} &
  0.86 &
  \multicolumn{1}{c|}{0.93} &
  0.84 &
  \textbf{0.94} \\
warpAR10P &
  0.33 &
  \multicolumn{1}{c|}{0.44} &
  0.56 &
  \multicolumn{1}{c|}{0.78} &
  0.72 &
  \multicolumn{1}{c|}{0.85} &
  0.70 &
  \multicolumn{1}{c|}{\textbf{0.95}} &
  0.75 &
  0.85 \\
warpPIE10P &
  0.85 &
  \multicolumn{1}{c|}{0.89} &
  0.95 &
  \multicolumn{1}{c|}{\textbf{1.00}} &
  0.98 &
  \multicolumn{1}{c|}{1.00} &
  1.00 &
  \multicolumn{1}{c|}{1.00} &
  1.00 &
  1.00 \\
Yale &
  0.14 &
  \multicolumn{1}{c|}{0.33} &
  0.25 &
  \multicolumn{1}{c|}{0.50} &
  0.39 &
  \multicolumn{1}{c|}{0.67} &
  0.37 &
  \multicolumn{1}{c|}{0.69} &
  0.53 &
  \textbf{0.70} \\
USPS &
  0.72 &
  \multicolumn{1}{c|}{0.65} &
  0.90 &
  \multicolumn{1}{c|}{0.90} &
  0.91 &
  \multicolumn{1}{c|}{0.92} &
  0.92 &
  \multicolumn{1}{c|}{\textbf{0.93}} &
  0.92 &
  0.93 \\
colon &
  0.70 &
  \multicolumn{1}{c|}{0.69} &
  0.69 &
  \multicolumn{1}{c|}{0.66} &
  \textbf{0.92} &
  \multicolumn{1}{c|}{0.82} &
  0.85 &
  \multicolumn{1}{c|}{0.74} &
  0.85 &
  0.88 \\
GLIOMA &
  \textbf{0.61} &
  \multicolumn{1}{c|}{0.25} &
  0.30 &
  \multicolumn{1}{c|}{0.30} &
  0.30 &
  \multicolumn{1}{c|}{0.38} &
  0.60 &
  \multicolumn{1}{c|}{0.41} &
  0.59 &
  0.25 \\
lung &
  0.39 &
  \multicolumn{1}{c|}{0.47} &
  0.67 &
  \multicolumn{1}{c|}{0.89} &
  0.81 &
  \multicolumn{1}{c|}{\textbf{0.95}} &
  0.71 &
  \multicolumn{1}{c|}{0.87} &
  0.90 &
  0.81 \\
lung\_small &
  0.49 &
  \multicolumn{1}{c|}{0.57} &
  0.76 &
  \multicolumn{1}{c|}{0.79} &
  0.82 &
  \multicolumn{1}{c|}{0.68} &
  0.79 &
  \multicolumn{1}{c|}{0.75} &
  0.82 &
  \textbf{0.93} \\
lymphoma &
  0.22 &
  \multicolumn{1}{c|}{0.50} &
  0.58 &
  \multicolumn{1}{c|}{0.96} &
  0.78 &
  \multicolumn{1}{c|}{0.87} &
  0.90 &
  \multicolumn{1}{c|}{0.82} &
  0.81 &
  \textbf{0.98} \\
GISETTE &
  0.50 &
  \multicolumn{1}{c|}{0.49} &
  0.48 &
  \multicolumn{1}{c|}{0.47} &
  0.51 &
  \multicolumn{1}{c|}{\textbf{0.52}} &
  0.47 &
  \multicolumn{1}{c|}{0.50} &
  0.49 &
  0.50 \\
Isolet &
  0.32 &
  \multicolumn{1}{c|}{0.51} &
  0.74 &
  \multicolumn{1}{c|}{0.78} &
  0.81 &
  \multicolumn{1}{c|}{0.82} &
  0.88 &
  \multicolumn{1}{c|}{0.83} &
  0.89 &
  \textbf{0.89} \\
MADELON &
  0.59 &
  \multicolumn{1}{c|}{\textbf{0.59}} &
  0.58 &
  \multicolumn{1}{c|}{0.56} &
  0.55 &
  \multicolumn{1}{c|}{0.57} &
  0.54 &
  \multicolumn{1}{c|}{0.57} &
  0.57 &
  0.57 \\ \hline\hline
  \# bests&
  5 &
  \multicolumn{1}{c|}{11} &
  3 &
  \multicolumn{1}{c|}{13} &
  2 &
  \multicolumn{1}{c|}{14} &
  5 &
  \multicolumn{1}{c|}{11} &
  2 &
  14 \\ \hline
\end{tabular}
}
\end{minipage}
\hspace{0.7in}
\begin{minipage}[t]{0.4\linewidth}
\centering
\scalebox{0.52}{
\begin{tabular}{c|cccccccccc}
\hline
 &
  \multicolumn{10}{c}{\textbf{KNN}} \\ \hline
 &
  \multicolumn{2}{c|}{\textbf{10}} &
  \multicolumn{2}{c|}{\textbf{50}} &
  \multicolumn{2}{c|}{\textbf{100}} &
  \multicolumn{2}{c|}{\textbf{150}} &
  \multicolumn{2}{c}{\textbf{200}} \\ \cline{2-11} 
 &
  \textbf{$\textrm{Inf-FS}_U$} &
  \multicolumn{1}{c|}{\textbf{TFS}} &
  \textbf{$\textrm{Inf-FS}_U$} &
  \multicolumn{1}{c|}{\textbf{TFS}} &
  \textbf{$\textrm{Inf-FS}_U$} &
  \multicolumn{1}{c|}{\textbf{TFS}} &
  \textbf{$\textrm{Inf-FS}_U$} &
  \multicolumn{1}{c|}{\textbf{TFS}} &
  \textbf{$\textrm{Inf-FS}_U$} &
  \textbf{TFS} \\ \hline
PCMAC &
  0.52 &
  \multicolumn{1}{c|}{0.53} &
  0.57 &
  \multicolumn{1}{c|}{0.61} &
  0.61 &
  \multicolumn{1}{c|}{0.62} &
  0.59 &
  \multicolumn{1}{c|}{\textbf{0.63}} &
  0.60 &
  0.62 \\
RELATHE &
  0.46 &
  \multicolumn{1}{c|}{0.46} &
  0.50 &
  \multicolumn{1}{c|}{\textbf{0.57}} &
  0.48 &
  \multicolumn{1}{c|}{0.49} &
  0.46 &
  \multicolumn{1}{c|}{0.45} &
  0.48 &
  0.48 \\
COIL20 &
  0.70 &
  \multicolumn{1}{c|}{0.82} &
  0.86 &
  \multicolumn{1}{c|}{0.93} &
  0.93 &
  \multicolumn{1}{c|}{0.93} &
  0.96 &
  \multicolumn{1}{c|}{0.94} &
  \textbf{0.97} &
  0.93 \\
ORL &
  0.38 &
  \multicolumn{1}{c|}{0.52} &
  0.52 &
  \multicolumn{1}{c|}{\textbf{0.77}} &
  0.62 &
  \multicolumn{1}{c|}{0.70} &
  0.73 &
  \multicolumn{1}{c|}{0.71} &
  0.72 &
  0.77 \\
warpAR10P &
  0.36 &
  \multicolumn{1}{c|}{0.30} &
  0.36 &
  \multicolumn{1}{c|}{\textbf{0.51}} &
  0.43 &
  \multicolumn{1}{c|}{0.46} &
  0.32 &
  \multicolumn{1}{c|}{0.38} &
  0.42 &
  0.48 \\
warpPIE10P &
  0.83 &
  \multicolumn{1}{c|}{0.72} &
  0.86 &
  \multicolumn{1}{c|}{0.91} &
  0.92 &
  \multicolumn{1}{c|}{\textbf{0.97}} &
  0.89 &
  \multicolumn{1}{c|}{0.92} &
  0.89 &
  0.95 \\
Yale &
  0.14 &
  \multicolumn{1}{c|}{0.42} &
  0.28 &
  \multicolumn{1}{c|}{0.41} &
  0.26 &
  \multicolumn{1}{c|}{0.42} &
  0.43 &
  \multicolumn{1}{c|}{0.38} &
  0.35 &
  \textbf{0.49} \\
USPS &
  0.78 &
  \multicolumn{1}{c|}{0.77} &
  0.94 &
  \multicolumn{1}{c|}{0.94} &
  \textbf{0.96} &
  \multicolumn{1}{c|}{0.95} &
  0.96 &
  \multicolumn{1}{c|}{0.95} &
  0.95 &
  0.95 \\
colon &
  0.77 &
  \multicolumn{1}{c|}{0.82} &
  0.89 &
  \multicolumn{1}{c|}{0.77} &
  0.89 &
  \multicolumn{1}{c|}{0.85} &
  \textbf{1.00} &
  \multicolumn{1}{c|}{0.70} &
  0.85 &
  0.77 \\
GLIOMA &
  0.24 &
  \multicolumn{1}{c|}{0.10} &
  0.40 &
  \multicolumn{1}{c|}{0.40} &
  0.42 &
  \multicolumn{1}{c|}{\textbf{0.62}} &
  0.52 &
  \multicolumn{1}{c|}{0.62} &
  0.52 &
  0.62 \\
lung &
  0.33 &
  \multicolumn{1}{c|}{0.51} &
  0.65 &
  \multicolumn{1}{c|}{\textbf{0.79}} &
  0.71 &
  \multicolumn{1}{c|}{0.65} &
  0.72 &
  \multicolumn{1}{c|}{0.68} &
  0.78 &
  0.79 \\
lung\_small &
  0.57 &
  \multicolumn{1}{c|}{0.61} &
  0.80 &
  \multicolumn{1}{c|}{0.87} &
  0.82 &
  \multicolumn{1}{c|}{0.90} &
  \textbf{0.93} &
  \multicolumn{1}{c|}{0.90} &
  0.90 &
  0.76 \\
lymphoma &
  0.44 &
  \multicolumn{1}{c|}{0.50} &
  0.60 &
  \multicolumn{1}{c|}{0.74} &
  0.69 &
  \multicolumn{1}{c|}{0.69} &
  \textbf{0.76} &
  \multicolumn{1}{c|}{0.75} &
  0.69 &
  0.74 \\
GISETTE &
  0.49 &
  \multicolumn{1}{c|}{0.51} &
  0.52 &
  \multicolumn{1}{c|}{\textbf{0.54}} &
  0.50 &
  \multicolumn{1}{c|}{0.51} &
  0.50 &
  \multicolumn{1}{c|}{0.53} &
  0.50 &
  0.49 \\
Isolet &
  0.32 &
  \multicolumn{1}{c|}{0.49} &
  0.72 &
  \multicolumn{1}{c|}{0.73} &
  0.78 &
  \multicolumn{1}{c|}{0.78} &
  \textbf{0.83} &
  \multicolumn{1}{c|}{0.81} &
  0.82 &
  0.83 \\
MADELON &
  0.61 &
  \multicolumn{1}{c|}{\textbf{0.78}} &
  0.58 &
  \multicolumn{1}{c|}{0.74} &
  0.64 &
  \multicolumn{1}{c|}{0.66} &
  0.62 &
  \multicolumn{1}{c|}{0.64} &
  0.57 &
  0.63 \\ \hline\hline
  \# bests &
  4 &
  \multicolumn{1}{c|}{12} &
  1 &
  \multicolumn{1}{c|}{15} &
  3 &
  \multicolumn{1}{c|}{13} &
  10 &
  \multicolumn{1}{c|}{6} &
  4 &
  12 \\ \hline
\end{tabular}
}
\end{minipage}

\vspace{0.1in}
\centering
\scalebox{0.52}{
\begin{tabular}{c|cccccccccc}
\hline
 &
  \multicolumn{10}{c}{\textbf{Decision Tree}} \\ \hline
 &
  \multicolumn{2}{c|}{\textbf{10}} &
  \multicolumn{2}{c|}{\textbf{50}} &
  \multicolumn{2}{c|}{\textbf{100}} &
  \multicolumn{2}{c|}{\textbf{150}} &
  \multicolumn{2}{c}{\textbf{200}} \\ \cline{2-11} 
 &
  \textbf{$\textrm{Inf-FS}_U$} &
  \multicolumn{1}{c|}{\textbf{TFS}} &
  \textbf{$\textrm{Inf-FS}_U$} &
  \multicolumn{1}{c|}{\textbf{TFS}} &
  \textbf{$\textrm{Inf-FS}_U$} &
  \multicolumn{1}{c|}{\textbf{TFS}} &
  \textbf{$\textrm{Inf-FS}_U$} &
  \multicolumn{1}{c|}{\textbf{TFS}} &
  \textbf{$\textrm{Inf-FS}_U$} &
  \textbf{TFS} \\ \hline
PCMAC &
  0.53 &
  \multicolumn{1}{c|}{0.50} &
  0.56 &
  \multicolumn{1}{c|}{0.69} &
  0.58 &
  \multicolumn{1}{c|}{0.71} &
  0.57 &
  \multicolumn{1}{c|}{0.68} &
  0.60 &
  \textbf{0.73} \\
RELATHE &
  0.49 &
  \multicolumn{1}{c|}{0.50} &
  0.51 &
  \multicolumn{1}{c|}{\textbf{0.51}} &
  0.49 &
  \multicolumn{1}{c|}{0.42} &
  0.48 &
  \multicolumn{1}{c|}{0.51} &
  0.48 &
  0.51 \\
COIL20 &
  0.68 &
  \multicolumn{1}{c|}{0.81} &
  0.83 &
  \multicolumn{1}{c|}{0.89} &
  0.85 &
  \multicolumn{1}{c|}{\textbf{0.90}} &
  0.89 &
  \multicolumn{1}{c|}{0.90} &
  0.90 &
  0.90 \\
ORL &
  0.36 &
  \multicolumn{1}{c|}{0.39} &
  0.42 &
  \multicolumn{1}{c|}{0.48} &
  0.49 &
  \multicolumn{1}{c|}{0.54} &
  0.59 &
  \multicolumn{1}{c|}{0.61} &
  0.49 &
  \textbf{0.62} \\
warpAR10P &
  0.37 &
  \multicolumn{1}{c|}{0.33} &
  0.46 &
  \multicolumn{1}{c|}{0.59} &
  0.55 &
  \multicolumn{1}{c|}{0.59} &
  0.41 &
  \multicolumn{1}{c|}{0.64} &
  0.68 &
  \textbf{0.80} \\
warpPIE10P &
  0.74 &
  \multicolumn{1}{c|}{0.74} &
  0.80 &
  \multicolumn{1}{c|}{0.73} &
  0.77 &
  \multicolumn{1}{c|}{0.85} &
  0.76 &
  \multicolumn{1}{c|}{\textbf{0.87}} &
  0.76 &
  0.81 \\
Yale &
  0.17 &
  \multicolumn{1}{c|}{0.31} &
  0.26 &
  \multicolumn{1}{c|}{0.34} &
  0.39 &
  \multicolumn{1}{c|}{0.42} &
  0.50 &
  \multicolumn{1}{c|}{0.43} &
  0.43 &
  \textbf{0.52} \\
USPS &
  0.73 &
  \multicolumn{1}{c|}{0.72} &
  0.84 &
  \multicolumn{1}{c|}{0.85} &
  0.85 &
  \multicolumn{1}{c|}{0.86} &
  0.86 &
  \multicolumn{1}{c|}{0.86} &
  \textbf{0.88} &
  0.87 \\
colon &
  0.61 &
  \multicolumn{1}{c|}{0.64} &
  0.82 &
  \multicolumn{1}{c|}{0.74} &
  0.76 &
  \multicolumn{1}{c|}{\textbf{0.92}} &
  0.89 &
  \multicolumn{1}{c|}{0.83} &
  0.85 &
  0.83 \\
GLIOMA &
  0.34 &
  \multicolumn{1}{c|}{\textbf{0.61}} &
  0.36 &
  \multicolumn{1}{c|}{0.31} &
  0.35 &
  \multicolumn{1}{c|}{0.44} &
  0.38 &
  \multicolumn{1}{c|}{0.31} &
  0.31 &
  0.44 \\
lung &
  0.44 &
  \multicolumn{1}{c|}{0.70} &
  0.75 &
  \multicolumn{1}{c|}{0.71} &
  0.87 &
  \multicolumn{1}{c|}{0.70} &
  \textbf{0.90} &
  \multicolumn{1}{c|}{0.73} &
  0.71 &
  0.79 \\
lung\_small &
  0.46 &
  \multicolumn{1}{c|}{0.42} &
  0.58 &
  \multicolumn{1}{c|}{\textbf{0.63}} &
  0.47 &
  \multicolumn{1}{c|}{0.57} &
  0.52 &
  \multicolumn{1}{c|}{0.63} &
  0.52 &
  0.49 \\
lymphoma &
  0.20 &
  \multicolumn{1}{c|}{\textbf{0.69}} &
  0.45 &
  \multicolumn{1}{c|}{0.55} &
  0.45 &
  \multicolumn{1}{c|}{0.44} &
  0.63 &
  \multicolumn{1}{c|}{0.60} &
  0.51 &
  0.62 \\
GISETTE &
  0.52 &
  \multicolumn{1}{c|}{0.50} &
  0.44 &
  \multicolumn{1}{c|}{\textbf{0.52}} &
  0.48 &
  \multicolumn{1}{c|}{0.47} &
  0.50 &
  \multicolumn{1}{c|}{0.49} &
  0.49 &
  0.48 \\
Isolet &
  0.27 &
  \multicolumn{1}{c|}{0.43} &
  0.69 &
  \multicolumn{1}{c|}{0.67} &
  0.73 &
  \multicolumn{1}{c|}{0.71} &
  0.74 &
  \multicolumn{1}{c|}{0.73} &
  \textbf{0.78} &
  0.73 \\
MADELON &
  0.58 &
  \multicolumn{1}{c|}{0.66} &
  0.70 &
  \multicolumn{1}{c|}{\textbf{0.81}} &
  0.78 &
  \multicolumn{1}{c|}{0.79} &
  0.75 &
  \multicolumn{1}{c|}{0.77} &
  0.74 &
  0.77 \\ \hline\hline
  \# bests &
  6 &
  \multicolumn{1}{c|}{10} &
  5 &
  \multicolumn{1}{c|}{11} &
  5 &
  \multicolumn{1}{c|}{11} &
  7 &
  \multicolumn{1}{c|}{9} &
  5 &
  11 \\ \hline
\end{tabular}
}

\end{table*}

TFS combined with LinearSVM classifier produces higher Balanced Accuracy scores in $14$ out of $16$ datasets (i.e. $87.5\%$ of cases), while $\textrm{Inf-FS}_U$ combined with the same classifier has a higher Balanced Accuracy scores only in $2$ out of $16$ datasets (i.e. $12.5\%$ of cases). Considering only the scenarios where TFS is the winning feature selection schema, we notice that in $1$ case, the optimal cardinality is equal to $10$, in $2$ cases, the optimal cardinality is equal to $50$ and $100$, in three cases the optimal cardinality is equal to $150$, and in $6$ cases the optimal cardinality is equal to $200$. When the $\textrm{Inf-FS}_U$ is used as feature selection algorithm and LinearSVM as classifier, the cross-datasets average Balanced Accuracy score grows from a value equal to $0.49$ at cardinality $10$ to a value of $0.75$ at cardinality $200$ with an increase of $26\%$. When the TFS is used as feature selection algorithm and LinearSVM as classifier, cross-datasets average Balanced Accuracy score grows from a value equal to $0.52$ at cardinality $10$ to a value of $0.78$ at cardinality $200$ with an increase of $26\%$. Finally, we notice that when the $\textrm{Inf-FS}_U$ is used as feature selection algorithm and LinearSVM as classifier, the average maximum drawdown ratio is equal to $31\%$. Using TFS as feature selection algorithm, instead, the average maximum drawdown ratio is equal to $28\%$.

TFS combined with KNN classifier produces higher Balanced Accuracy scores in $10$ out of $16$ datasets (i.e. $62.5\%$ of cases), while $\textrm{Inf-FS}_U$ combined with the same classifier has higher Balanced Accuracy scores in $6$ out of $16$ datasets (i.e. $37.5\%$ of cases). Considering only the scenarios where TFS is the winning feature selection schema, we notice that in $1$ case, the optimal cardinality is equal to $10$, $150$ and $200$, in $5$ cases, the optimal cardinality is equal to $50$, in two cases the optimal cardinality is equal to $100$. When the $\textrm{Inf-FS}_U$ is used as feature selection algorithm and KNN as classifier, the cross-datasets average Balanced Accuracy score grows from a value equal to $0.46$ at cardinality $10$ to a value of $0.69$ at cardinality $200$ with an increase of $23\%$. When TFS is used as feature selection algorithm and KNN as classifier, the cross-datasets average Balanced Accuracy score grows from a value equal to $0.55$ at cardinality $10$ to a value of $0.71$ at cardinality $200$ with an increase of $16\%$. Finally, we notice that when $\textrm{Inf-FS}_U$ is used as feature selection algorithm and KNN as classifier, the average maximum drawdown ratio is equal to $23\%$. Using TFS as feature selection algorithm, instead, the average maximum drawdown ratio is equal to $21\%$.

TFS combined with Decision Tree classifier produces higher Balanced Accuracy scores in $13$ out of $16$ datasets (i.e. $81.25\%$ of cases), while $\textrm{Inf-FS}_U$ combined with the same classifier produces higher Balanced Accuracy scores in $3$ out of $16$ datasets (i.e. $18.75\%$ of cases). Considering only the scenarios where TFS is the winning feature selection schema, we notice that in $2$ cases, the optimal cardinality is equal to $10$ and $100$, in $4$ cases, the optimal cardinality is equal to $50$ and $200$, in only one case the optimal cardinality is equal to $150$.  When the $\textrm{Inf-FS}_U$ is used as feature selection algorithm and Decision Tree as classifier, the cross-datasets average Balanced Accuracy score grows from a value equal to $0.47$ at cardinality $10$, to a value of $0.63$ at cardinality $200$ with an increase of $16\%$. When TFS is used as feature selection algorithm and Decision Tree as classifier, the cross-datasets average Balanced Accuracy score grows from a value equals to $0.56$ at cardinality $10$, to a value of $0.68$ at cardinality $200$ with an increase of $12\%$. Finally, we notice that, when $\textrm{Inf-FS}_U$ is used as feature selection algorithm and Decision Tree as classifier, the average maximum drawdown ratio is equal to $22\%$. Using TFS as feature selection algorithm, instead, the average maximum drawdown ratio is equal to $20\%$.

Considering the cross-datasets average Balanced Accuracy scores discussed earlier in this Section, we conclude that, independently from the chosen classifier, TFS generally allows to select more informative features, guaranteeing higher learning performances. Both the  cross-datasets average Balanced Accuracy score percentage increase and the cross-datasets average maximum drawdown ratio are lower when TFS is chosen as feature selection schema, further certifying an higher stability and an ability to choose higher quality features.

\begin{table}[h]
\centering
\caption{Out-of-sample Balanced Accuracy scores obtained by LinearSVM, KNN and Decision Tree classifier on the raw datasets (i.e. the datasets containing all the original features). We boldly highlight the entries where a TFS improves the classifier's performance. We do not highlight the entries where a classifier performs better on the raw dataset (i.e. where feature selection algorithms $\textrm{Inf-FS}_U$ and TFS are not effective). The $^*$ symbol highlights the scenarios where the optimal feature selection schema is $\textrm{Inf-FS}_U$ but TFS, in combination with the classifier, still outperforms the classifier on the raw dataset. The $^\dagger$ symbol highlights the scenarios where the optimal feature selection schema is $\textrm{Inf-FS}_U$ while TFS, in combination with the classifier, cannot outperform the classifier on the raw dataset.}
\label{tab:Full_dataset_results}
\vspace{0.1in}
\scalebox{0.7}{%
\begin{tabular}{c|c|c|c}
\hline
            & \textbf{LinearSVM} & \textbf{KNN}  & \textbf{Decision Tree}   \\ \hline
PCMAC       & 0.83     & 0.71 & 0.90  \\
RELATHE     & 0.84     & 0.77 & 0.85 \\
COIL20      & 0.97      & \;0.96$^\dagger$          & \textbf{0.87}          \\
ORL         & 0.97      & 0.82 & 0.68 \\
warpAR10P   & 1.00      & 0.53 & \textbf{0.66}          \\
warpPIE10P  & \textbf{1.00}               & \textbf{0.84}          & \textbf{0.84}          \\
Yale        & 0.82               & \textbf{0.46}          & \textbf{0.44}          \\
USPS        & \textbf{0.93}               & \;0.95$^*$          & \;0.87$^*$          \\
colon       & \;0.80$^*$               & \;0.73$^*$          & \textbf{0.80}           \\
GLIOMA      & \;0.56$^\dagger$               & 0.78 & \textbf{0.44}          \\
lung        & \textbf{0.93}               & \textbf{0.76}          & \;0.66$^*$          \\
lung\_small & \textbf{0.79}               & \;0.76$^*$          & \textbf{0.63}          \\
lymphoma    & \textbf{0.93}               & \;0.69$^*$          & \textbf{0.61}          \\
GISETTE     & 0.98     & 0.96 & 0.92 \\
Isolet      & 0.93     & 0.84 & 0.79 \\
MADELON     & \textbf{0.58}               & \textbf{0.51}          & \textbf{0.74}          \\ \hline
\end{tabular}%
}
\end{table}

The power of TFS can be further investigated by comparing results obtained by applying the three classification algorithms on the raw datasets (i.e. the datasets containing all the original features) against the best ones obtained by applying the same classifiers combined with the novel feature classification technique presented in this paper. Results are reported in Table \ref{tab:Full_dataset_results}. When LinearSVM is chosen as a classifier, feature selection turns out to be beneficial on $9$ out of $16$ datasets (i.e. $56.25\%$ of cases). In $7$ cases, TFS is the optimal feature classification approach; in the case of the ‘colon' dataset, even if $\textrm{Inf-FS}_U$ is the optimal feature selection approach, results achieved using TFS are still better than the ones obtained on the raw dataset; in the case of ‘GLIOMA' dataset, $\textrm{Inf-FS}_U$ is the optimal feature selection approach and results achieved using TFS are lower than the ones obtained on the raw dataset. When KNN is chosen as a classifier, feature selection is beneficial on $9$ out of $16$ datasets (i.e. $56.25\%$ of cases). In $4$ cases, TFS is the optimal feature classification approach; in the case of ‘USPS', ‘colon', ‘lung\_small' and ‘lymphoma' dataset, even if $\textrm{Inf-FS}_U$ is the optimal feature selection approach, results achieved using TFS are still better than the ones obtained on the raw dataset; in the case of ‘COIL20' dataset, $\textrm{Inf-FS}_U$ is the optimal feature selection approach and results achieved using TFS are lower than the ones obtained on the raw dataset. Finally, when Decision Tree is chosen as a classifier, feature selection is beneficial on $11$ out of $16$ datasets (i.e. $68.75\%$ of cases). In $9$ cases, TFS is the optimal feature classification approach; in the case of ‘USPS' and ‘colon' dataset, even if $\textrm{Inf-FS}_U$ is the optimal feature selection approach, results achieved using TFS are still better than the ones obtained on the raw dataset. \\
Looking at the results reported above, we notice that the application domains where TFS is more compelling are the ones where the tabular format is the natural data format (i.e. biological data and artificial data). This finding is not unexpected. Application domains such as text, face images and spoken letters data would require a more complex data pre-processing pipeline (e.g. encoding) and specific deep learning-based classification algorithms (e.g. convolutional and recurrent neural networks). A deeper analysis of this aspect is left for the upcoming TFS-centered research.

The statistical significance of results discussed earlier in this Section is assessed in Appendix \ref{Appendix_7}. Looking at the entries of Tables \ref{tab:local_bests_results} where TFS defines a new state-of-the-art, we state that (i) when LinearSVM is used as classifier, TFS is statistically different from $\textrm{Inf-FS}_U$ in $8$ out of $14$ cases ($57\%$); (ii) when KNN is used as classifier, TFS is statistically different from $\textrm{Inf-FS}_U$ in $3$ out of $10$ cases ($30\%$); (iii) when Decision Tree is used as classifier, TFS is statistically different from $\textrm{Inf-FS}_U$ in $4$ out of $13$ cases ($31\%$). There is only one dataset (i.e. ‘GISETTE’) where TFS is statistically different from $\textrm{Inf-FS}_U$ independently from the classifier: in all the other cases, results are dependent on the choice of the classifier.

\section{Conclusions} \label{sec:Conclusions}
In this work, we combine the power of state-of-the-art IFNs, and instruments from network science to develop a novel unsupervised, graph-based filter method for feature selection. Features are represented as nodes in a TMFG, and their relevance is assessed by studying their relative position inside the network. Exploiting topological properties of the network used to represent meaningful interactions among features, we propose a physics-informed (i.e. we use instruments from complexity science) feature selection model that is highly flexible, computationally cheap, fully explainable and remarkably intuitive. To prove the effectiveness of the proposed methodology, we test it against the current state-of-the-art counterpart (i.e. $\textrm{Inf-FS}_U$) on $16$ benchmark datasets belonging to different applicative domains. Employing a Linear Support Vector classifier, a $k$-Nearest Neighbors classifier and a Decision Tree classifier, we show how our algorithm achieves top performances on most benchmark datasets, redefining the current state-of-the-art on a significant number of them. The proposed methodology demonstrates effectiveness in conditions where the amount of training data largely varies. Compared to its main alternative, TFS has a lower computational complexity and provides a much more intuitive overview of the feature selection process. Thanks to the possibility of studying the relative position of nodes in the network in many different ways (i.e. choosing different centrality measures or defining new ones), TFS is highly versatile and fully adaptable to input data. It is worth noting that the current work is a foundational one. It presents three aspects that are unsatisfactory and we plan to cover in the future: (i) the need to explicitly specify the cardinality of the subset of relevant features is limiting and requires an a priori knowledge of the applicative domain or, at least, an extended search for the optimal realization of this hyper-parameter; (ii) the usage of classic correlation measures in the TMFG's building process prevents from the possibility to handle problems with mixed type of features (continuous-categorical, categorical-categorical); (iii) TMFG is non-differentiable and this prevents from a direct integration with advanced Deep Learning-based architectures. More generally, this study points to many future directions spanning from the development of data-centred measures to assess features relevance, to the possibility of replicating the potential of this method and inferring it through automated learning techniques. The first steps have been taken in the latter research direction by introducing a new and potentially groundbreaking family of Neural Networks called Homological Neural Networks \cite{wang2023homological}. 

\section*{Acknowledgements}

Both the authors acknowledge Agne Scalchi, Silvia Bartolucci and Paolo Barucca for for the fruitful discussions on foundational topics related to this work. Both the authors acknowledge the ICML TAG-ML 2023 workshop's organising committee and the reviewers for the useful comments that improved the quality of the paper. The author, T.A., acknowledges the financial support from ESRC (ES/K002309/1), EPSRC (EP/P031730/1) and EC (H2020-ICT-2018-2 825215).

\nocite{langley00}

\bibliography{icml2023_tag/bibliography}
\bibliographystyle{icml2023}

\newpage
\appendix
\onecolumn

\section{Benchmark Datasets}\label{Appendix_1}

To prove TFS' effectiveness, we extensively test it on $16$ benchmark datasets (all in a tabular format) belonging to different application domains. For each dataset, Table \ref{tab:Datasets_Description} reports, respectively, the name, the application domain, the reference paper (or website), the downloading source, the number of features, the number of samples, the number of classes and the split dynamics (see Appendix \ref{Appendix_2} for further details on this last point).

\begin{table}[H]
\centering
\caption{Benchmark datasets used to compare feature selection algorithms considered in the current work. The order of appearance is inherited from \cite{ASUDatasets}.}
\label{tab:Datasets_Description}
\vspace{0.1in}
\scalebox{0.7}{%
\begin{tabular}{@{}cccccccc@{}}
\toprule
\textbf{Name} &
  \textbf{Category} &
  \textbf{Reference} &
  \textbf{Source} &
  \textbf{\# Features} &
  \textbf{\# Samples} &
  \textbf{Classes} &
  \textbf{Split Provided} \\ \midrule
PCMAC       & Text Data              & \cite{lang1995newsweeder}   & \cite{ASUDatasets} & 3289 & 1943 & binary      & false \\
RELATHE     & Text Data              & \cite{lang1995newsweeder}   & \cite{ASUDatasets} & 4322 & 1427 & binary      & false \\
COIL20      & Face Images Data         & \cite{nene1996columbia}     & \cite{ASUDatasets} & 1024 & 1440 & multi-class & false \\
ORL &
  Face Images Data &
  \cite{samaria1994parameterisation} &
  \cite{ASUDatasets} &
  1024 &
  400 &
  multi-class &
  false \\
warpAR10P   & Face Images Data        & \cite{li2018feature}        & \cite{ASUDatasets} & 2400 & 130  & multi-class & false \\
warpPIE10P  & Face Images Data       & \cite{sim2002cmu}           & \cite{ASUDatasets} & 2420 & 210  & multi-class & false \\
Yale &
  Face Images Data &
  \cite{belhumeur1997eigenfaces} &
  \cite{ASUDatasets} &
  1024 &
  165 &
  multi-class &
  false \\
USPS        & Hand Written Images Data & \cite{hull1994database}     & \cite{GitHubIFS}     & 256  & 9298 & multi-class & false \\
colon       & Biological Data         & \cite{alon1999broad}        & \cite{GitHubIFS}     & 2000 & 62   & binary      & false \\
GLIOMA      & Biological Data         & \cite{li2018feature}        & \cite{ASUDatasets} & 4434 & 50   & multi-class & false \\
lung        & Biological Data        & \cite{GitHubIFS}            & \cite{GitHubIFS}     & 3312 & 203  & multi-class & false \\
lung\_small & Biological Data        & \cite{GitHubIFS}            & \cite{GitHubIFS}     & 325  & 73   & multi-class & false \\
lymphoma    & Biological Data        & \cite{golub1999molecular}   & \cite{ASUDatasets} & 4026 & 96   & multi-class & false \\
GISETTE     & Digits Data  & \cite{guyon2007competitive} & \cite{Dua:2019}      & 5000 & 7000 & binary      & true  \\
Isolet &
  Spoken Letters Data &
  \cite{li2018feature} &
  \cite{ASUDatasets} &
  617 &
  1560 &
  multi-class &
  false \\
MADELON     & Artificial Data         & \cite{guyon2007competitive} & \cite{Dua:2019}      & 500  & 2600 & binary      & true  \\ \bottomrule
\end{tabular}%
}
\end{table}

We distinguish among $7$ different application domains (i.e. text data, face images data, hand written images data, biological data, digits data, spoken letters data and artificial data). Categories follow the taxonomy in \cite{li2018feature}. The average number of features is $2248$. The dataset with the lowest number of features is ‘USPS’ (i.e. $256$). The dataset with the largest number of features is ‘GISETTE’ (i.e. $5000$). The average number of samples is $1666$. The dataset with the lowest number of samples is ‘GLIOMA’ (i.e. $50$), while the one with the largest number of samples is ‘USPS’ (i.e. $9298$). $5$ of the considered datasets are binary, while $11$ are multi-class.

\newpage 

\section{Data Pipeline}\label{Appendix_2}

We design the data pre-processing pipeline as consisting of $3$ different steps: (i) data reading and format unification; (ii) training/validation/test splitting; (iii) constant features pruning. The first step allows us to read data and unify formats coming from different sources. The second step consists of the training/validation/test splitting. Depending on the source, training/test split could be provided or not. The two datasets ‘GISETTE’ and ‘MADELON’ come with a provided training/validation/test split. In both cases, the data source (i.e. \cite{Dua:2019}) does not provide test labels. Because of this, we use the validation set for testing. For all the other datasets, $70\%$ of the raw dataset is used as a training and validation set, while $30\%$ is used as a test set. We use a stratified splitting procedure to ensure that each set contains, for each target class, approximately the same percentage of samples as per in the raw dataset. In the third step, non-informative, constant covariates are detected on the training set and permanently removed from the training, validation and test set. 

Table \ref{tab:Pre_Processed_Data} reports datasets’ specifics after the preprocessing step. Looking at it, we remark that most considered datasets are not affected by the constant features filtering step. The only $4$ datasets which are reduced in the number of covariates are ‘PCMAC’ (with a reduction of $0.07\%$), ‘RELATHE’ (with a reduction of $0.03\%$), ‘GLIOMA’ (with a reduction of $0.03\%$) and ‘GISETTE’ (with a reduction of $0.9\%$). Table \ref{tab:Pre_Processed_Data} also reports the training and test dataset's labels' distributions. Datasets are generally balanced; the main exceptions are:

\begin{itemize}
    \item ‘USPS’: classes 1 and 2 are over-represented compared to the other classes.
    \item ‘colon’: class 1 is under-represented compared to the other class.
    \item ‘GLIOMA’: class 2 is under-represented compared to other classes.
    \item ‘lung’: class 1 is over-represented compared to the other classes.
    \item ‘lung\_small’: classes 1, 2, 3 and 5 are under-represented compared to the other classes.
    \item ‘lymphoma’: class 1 is over-represented compared to the other classes.
\end{itemize}

\begin{table}[H]
\caption{Datasets' specifics after the preprocessing step. For each benchmark dataset, we report the number of features, the number of samples and the labels' distribution for training and test data. The labels' distribution entry consists of a tuple for each class. The first element of the tuple represents the class itself, while the second represents the number of samples with that label.}
\label{tab:Pre_Processed_Data}
\vspace{0.1in}
\centering
\scalebox{0.55}{%
\begin{tabular}{|c|ccc|ccc|}
\hline
\textbf{Dataset} &
  \multicolumn{3}{c|}{\textbf{Training}} &
  \multicolumn{3}{c|}{\textbf{Test}} \\ \hline
 &
  \multicolumn{1}{c|}{\textbf{\# Features}} &
  \multicolumn{1}{c|}{\textbf{\# Samples}} &
  \textbf{Labels' Distribution} &
  \multicolumn{1}{c|}{\textbf{\# Features}} &
  \multicolumn{1}{c|}{\textbf{\# Samples}} &
  \textbf{Labels' Distribution} \\ \hline
PCMAC &
  \multicolumn{1}{c|}{3287} &
  \multicolumn{1}{c|}{1360} &
  (1, 687), (2, 673) &
  \multicolumn{1}{c|}{3287} &
  \multicolumn{1}{c|}{583} &
  (1, 295), (2, 288) \\ \hline
RELATHE &
  \multicolumn{1}{c|}{4321} &
  \multicolumn{1}{c|}{998} &
  (1, 545), (2, 453) &
  \multicolumn{1}{c|}{4321} &
  \multicolumn{1}{c|}{429} &
  (1, 234), (2, 195) \\ \hline
COIL20 &
  \multicolumn{1}{c|}{1024} &
  \multicolumn{1}{c|}{1008} &
  \begin{tabular}[c]{@{}c@{}}(1, 50), (2, 51), (3, 50), (4, 50), \\ (5, 50), (6, 50), (7, 51), (8, 50), \\ (9, 51),  (10, 50), (11, 51), (12, 50), \\ (13, 50),  (14, 51), (15, 50), (16, 50), \\ (17, 50),  (18, 51), (19, 51), (20, 51)\end{tabular} &
  \multicolumn{1}{c|}{1024} &
  \multicolumn{1}{c|}{432} &
  \begin{tabular}[c]{@{}c@{}}(1, 22), (2, 21), (3, 22), (4, 22), \\ (5, 22), (6, 22), (7, 21), (8, 22), \\ (9, 21), (10, 22), (11, 21), (12, 22), \\ (13, 22), (14, 21), (15, 22), (16, 22), \\ (17, 22), (18, 21), (19, 21), (20, 21)\end{tabular} \\ \hline
ORL &
  \multicolumn{1}{c|}{1024} &
  \multicolumn{1}{c|}{280} &
  \begin{tabular}[c]{@{}c@{}}(1, 7), (2, 7), (3, 7), (4, 7), (5, 7), \\ (6, 7), (7, 7), (8, 7), (9, 7), (10, 7),\\  (11, 7), (12, 7), (13, 7), (14, 7), (15, 7), \\ (16, 7), (17, 7), (18, 7), (19, 7), (20, 7), \\ (21, 7), (22, 7), (23, 7), (24, 7), (25, 7), \\ (26, 7), (27, 7), (28, 7), (29, 7), (30, 7), \\ (31, 7), (32, 7), (33, 7), (34, 7), (35, 7), \\ (36, 7), (37, 7), (38, 7), (39, 7), (40, 7)\end{tabular} &
  \multicolumn{1}{c|}{1024} &
  \multicolumn{1}{c|}{120} &
  \begin{tabular}[c]{@{}c@{}}(1, 3), (2, 3), (3, 3), (4, 3), (5, 3),\\  (6, 3), (7, 3), (8, 3), (9, 3), (10, 3), \\ (11, 3), (12, 3), (13, 3), (14, 3), (15, 3),\\  (16, 3), (17, 3), (18, 3), (19, 3), (20, 3), \\ (21, 3), (22, 3), (23, 3), (24, 3), (25, 3), \\ (26, 3), (27, 3), (28, 3), (29, 3), (30, 3), \\ (31, 3), (32, 3), (33, 3), (34, 3), (35, 3), \\ (36, 3), (37, 3), (38, 3), (39, 3), (40, 3)\end{tabular} \\ \hline
warpAR10P &
  \multicolumn{1}{c|}{2400} &
  \multicolumn{1}{c|}{91} &
  \begin{tabular}[c]{@{}c@{}}(1, 9), (2, 9), (3, 10), (4, 9), (5, 9), \\ (6, 9), (7, 9), (8, 9), (9, 9), (10, 9)\end{tabular} &
  \multicolumn{1}{c|}{2400} &
  \multicolumn{1}{c|}{39} &
  \begin{tabular}[c]{@{}c@{}}(1, 4), (2, 4), (3, 3), (4, 4), (5, 4), \\ (6, 4), (7, 4), (8, 4), (9, 4), (10, 4)\end{tabular} \\ \hline
warpPIE10P &
  \multicolumn{1}{c|}{2420} &
  \multicolumn{1}{c|}{147} &
  \begin{tabular}[c]{@{}c@{}}(1, 14), (2, 15), (3, 15), (4, 14), (5, 15),\\ (6, 14), (7, 15), (8, 15), (9, 15), (10, 15)\end{tabular} &
  \multicolumn{1}{c|}{2420} &
  \multicolumn{1}{c|}{63} &
  \begin{tabular}[c]{@{}c@{}}(1, 7), (2, 6), (3, 6), (4, 7), (5, 6), \\ (6, 7), (7, 6), (8, 6), (9, 6), (10, 6)\end{tabular} \\ \hline
Yale &
  \multicolumn{1}{c|}{1024} &
  \multicolumn{1}{c|}{115} &
  \begin{tabular}[c]{@{}c@{}}(1, 7), (2, 8), (3, 8), (4, 7), (5, 8), \\ (6, 7), (7, 8), (8, 8), (9, 8), (10, 8), \\ (11, 8), (12, 7), (13, 7), (14, 8), (15, 8)\end{tabular} &
  \multicolumn{1}{c|}{1024} &
  \multicolumn{1}{c|}{50} &
  \begin{tabular}[c]{@{}c@{}}(1, 4), (2, 3), (3, 3), (4, 4), (5, 3), \\ (6, 4), (7, 3), (8, 3), (9, 3), (10, 3), \\ (11, 3), (12, 4), (13, 4), (14, 3), (15, 3)\end{tabular} \\ \hline
USPS &
  \multicolumn{1}{c|}{256} &
  \multicolumn{1}{c|}{6508} &
  \begin{tabular}[c]{@{}c@{}}(1, 1087), (2, 888), (3, 650), (4, 577), (5, 596), \\ (6, 501), (7, 584), (8, 554), (9, 496), (10, 575)\end{tabular} &
  \multicolumn{1}{c|}{256} &
  \multicolumn{1}{c|}{2790} &
  \begin{tabular}[c]{@{}c@{}}(1, 466), (2, 381), (3, 279), (4, 247), (5, 256), \\ (6, 215), (7, 250), (8, 238), (9, 212), (10, 246)\end{tabular} \\ \hline
colon &
  \multicolumn{1}{c|}{2000} &
  \multicolumn{1}{c|}{43} &
  (-1, 28), (1, 15) &
  \multicolumn{1}{c|}{2000} &
  \multicolumn{1}{c|}{19} &
  (-1, 12), (1, 7) \\ \hline
GLIOMA &
  \multicolumn{1}{c|}{4433} &
  \multicolumn{1}{c|}{35} &
  (1, 10), (2, 5), (3, 10), (4, 10) &
  \multicolumn{1}{c|}{4433} &
  \multicolumn{1}{c|}{15} &
  (1, 4), (2, 2), (3, 4), (4, 5) \\ \hline
lung &
  \multicolumn{1}{c|}{3312} &
  \multicolumn{1}{c|}{142} &
  (1, 97), (2, 12), (3, 15), (4, 14), (5, 4) &
  \multicolumn{1}{c|}{3312} &
  \multicolumn{1}{c|}{61} &
  (1, 42), (2, 5), (3, 6), (4, 6), (5, 2) \\ \hline
lung\_small &
  \multicolumn{1}{c|}{325} &
  \multicolumn{1}{c|}{51} &
  \begin{tabular}[c]{@{}c@{}}(1, 4), (2, 3), (3, 4), (4, 11), \\ (5, 5), (6, 9), (7, 15)\end{tabular} &
  \multicolumn{1}{c|}{325} &
  \multicolumn{1}{c|}{22} &
  \begin{tabular}[c]{@{}c@{}}(1, 2), (2, 2), (3, 1), (4, 5), \\ (5, 2), (6, 4), (7, 6)\end{tabular} \\ \hline
lymphoma &
  \multicolumn{1}{c|}{4026} &
  \multicolumn{1}{c|}{67} &
  \begin{tabular}[c]{@{}c@{}}(1, 32), (2, 7), (3, 6), (4, 8), (5, 4), \\ (6, 4), (7, 3), (8, 1), (9, 2)\end{tabular} &
  \multicolumn{1}{c|}{4026} &
  \multicolumn{1}{c|}{29} &
  \begin{tabular}[c]{@{}c@{}}(1, 14), (2, 3), (3, 3), (4, 3), (5, 2), \\ (6, 2), (7, 1), (8, 1)\end{tabular} \\ \hline
GISETTE &
  \multicolumn{1}{c|}{4955} &
  \multicolumn{1}{c|}{6000} &
  (-1.0, 3000), (1.0, 3000) &
  \multicolumn{1}{c|}{4955} &
  \multicolumn{1}{c|}{1000} &
  (-1.0, 500), (1.0, 500) \\ \hline
Isolet &
  \multicolumn{1}{c|}{617} &
  \multicolumn{1}{c|}{1092} &
  \begin{tabular}[c]{@{}c@{}}(1, 42), (2, 42), (3, 42), (4, 42), (5, 42), \\ (6, 42), (7, 42), (8, 42), (9, 42), (10, 42), \\ (11, 42), (12, 42), (13, 42), (14, 42), (15, 42), \\ (16, 42), (17, 42), (18, 42), (19, 42), (20, 42), \\ (21, 42), (22, 42), (23, 42), (24, 42), (25, 42), (26, 42)\end{tabular} &
  \multicolumn{1}{c|}{617} &
  \multicolumn{1}{c|}{468} &
  \begin{tabular}[c]{@{}c@{}}(1, 18), (2, 18), (3, 18), (4, 18), (5, 18), \\ (6, 18), (7, 18), (8, 18), (9, 18), (10, 18), \\ (11, 18), (12, 18), (13, 18), (14, 18), (15, 18), \\ (16, 18), (17, 18), (18, 18), (19, 18), (20, 18), \\ (21, 18), (22, 18), (23, 18), (24, 18), (25, 18), (26, 18)\end{tabular} \\ \hline
MADELON &
  \multicolumn{1}{c|}{500} &
  \multicolumn{1}{c|}{2000} &
  (-1.0, 1000), (1.0, 1000) &
  \multicolumn{1}{c|}{500} &
  \multicolumn{1}{c|}{600} &
  (-1.0, 300), (1.0, 300) \\ \hline
\end{tabular}%
}
\end{table}

\newpage 

\section{Triangulated Maximally Filtered Graph}\label{Appendix_3}

The building process of the Triangulated Maximally Filtered Graph (TMFG) \cite{massara2017network} is based on a simple topological move that preserves the property of planarity: it adds one node to the centre of three-nodes cliques by using a score function that maximises the sum of the weights of the three edges connecting the existing vertices. This addition transforms three-nodes cliques (i.e. triangles) into four-nodes cliques (i.e. tetrahedrons) characterised by a chord that is not part of the clique but connects two nodes in the clique, forming two triangles and generating a chordal network \cite{christensen2018network}. The resulting network has $3n-6$ edges and is composed of three- and four-nodes cliques. TMFG has two relevant advantages: (i) it can be used to generate sparse probabilistic models as a form of topological regularization \cite{aste2022topological} and (ii) it is computationally efficient. On the other hand, the two main limitations of chordal networks are that (i) they may add unnecessary edges to satisfy the property of chordality and (ii) their building cost can vary based on the chosen optimization function.

\begin{algorithm}[H]
   \caption{TMFG built on the similarity matrix $\hat{\textbf{C}}$ to maximise the system's information flow.}\label{alg:TMFG_Algo}
\begin{algorithmic}
    \INPUT Similarity matrix $\hat{\textbf{C}}$ $\in \mathbb{R}^{n, n}$ from a set of observations $\{x_{1, 1}, \dots, x_{s, 1}\}, \{x_{1, 2}, \dots, x_{s, 2}\} \dots \{x_{1, n}, \dots, x_{s, n}\}$.
    \OUTPUT Sparse adjacency matrix \textbf{\textit{A}} describing the TMFG. 

    \hspace{0.15in}
    
    \FUNCTION{MaximumGain ($\hat{\textbf{C}}$, $\mathcal{V}$, $t$)}
    \STATE Initialize a vector of zeros $g \in \mathbb{R}^{1 \times n}$;
    \FOR{$j \in t$}
        \FOR{$v \notin \mathcal{V}$}
            \STATE $\hat{\textbf{C}}_{v,j} = 0$;
        \ENDFOR
        \STATE $g = g \oplus \hat{\textbf{C}}_{v,j}$;
    \ENDFOR
    \STATE \textbf{return} $\max{\{g\}}$.
    \ENDFUNCTION

    \hspace{0.15in}

    \STATE Initialize four empty sets: $\mathcal{C}$ (cliques), $\mathcal{T}$ (triangles), $\mathcal{S}$ (separators) and $\mathcal{V}$ (vertices);
    \STATE Initialize an adjacency matrix $\textbf{\textit{A}} \in \mathbb{R}^{n, n}$ with all zeros;
    \STATE $\mathcal{C}_1 \leftarrow$ tetrahedron, $\{v_1, v_2, v_3, v_4\}$, obtained choosing the $4$ entries of $\hat{\textbf{C}}$ maximising the similarity among features;
    \STATE $\mathcal{T} \leftarrow$ the four triangular faces in $\mathcal{C}_1: \{v_1, v_2, v_3\}, \{v_1, v_2, v_4\}, \{v_1, v_3, v_4\}, \{v_2, v_3, v_4\}$;
    \STATE $\mathcal{V} \leftarrow$ Assign to $\mathcal{V}$ the remaining $n-4$ vertices not in $\mathcal{C}_1$;
    \WHILE{$\mathcal{V}$ is not empty}
        \STATE Find the combination of $\{v_a, v_b, v_c\} \in \mathcal{T}$ (i.e. $t$) and $v_d \in \mathcal{V}$ which maximises MaximumGain($\hat{\textbf{C}}$, $\mathcal{V}$, $t$); 
        \STATE /* $\{v_a, v_b, v_c, v_d\}$ is a new 4-clique $\mathcal{C}$, $\{v_a, v_b, v_c\}$ becomes a separator $\mathcal{S}$, three new triangular faces, $\{v_a, v_b, v_d\}$, $\{v_a, v_c, v_d\}$ and $\{v_b, v_c, v_d\}$ are created */.
        \STATE Remove $v_d$ from $\mathcal{V}$;
        \STATE Remove $\{v_a, v_b, v_c\}$ from $\mathcal{T}$;
        \STATE Add $\{v_a, v_b, v_d\}$, $\{v_a, v_c, v_d\}$ and $\{v_b, v_c, v_d\}$ to $\mathcal{T}$;
    \ENDWHILE
    \STATE For each pair of nodes $i, j$ in $\mathcal{C}$, set $\textbf{\textit{A}}_{i,j} = 1$;
    \STATE \textbf{return} $\textbf{\textit{A}}$.
\end{algorithmic}
\end{algorithm}

\newpage

\section{Classification algorithms} \label{Appendix_4}
As reported in Section \ref{sec:Experiments}, the meaningfulness of the features' subsets chosen by $\textrm{Inf-FS}_U$ and TFS is evaluated based on the performance achieved by three classification algorithms: (i) Linear Support Vector Classifier; (ii) \textit{k}-Nearest Neighbors Classifier; (iii) Decision Tree Classifier. For all of them, we use the implementation provided by the ‘scikit-learn’ Python package \cite{scikit-learn}. The interested reader is referred to the following links for the implementations:

\begin{itemize}
    \item Linear Support Vector Classifier: \url{https://github.com/scikit-learn/scikit-learn/blob/98cf537f5/sklearn/svm/_classes.py#L14}
    \item \textit{k}-Nearest Neighbors Classifier: \url{https://github.com/scikit-learn/scikit-learn/blob/98cf537f5/sklearn/neighbors/_classification.py#L24}
    \item Decision Tree Classifier: \url{https://github.com/scikit-learn/scikit-learn/blob/98cf537f5/sklearn/tree/_classes.py#L595}
\end{itemize}

For the current research work we do not significantly change the models' default hyper-parameters. The first adjustment is performed on the Linear Support Vector Classifier's \verb|max_iter| parameter, which is set to $50000$. For Linear Support Vector Classifier and Decision Tree Classifier, the \verb|random_seed| parameter is set to $0$.

\newpage

\section{Optimal hyper-parameters configurations} \label{Appendix_5}
For each benchmark dataset, subset's cardinality and classification algorithm described in Section \ref{sec:Experiments}, Tables \ref{tab:Hyperopt_IFS} and \ref{tab:Hyperopt_TFS}, report optimal hyper-parameters configurations. 

It is worth mentioning that after the feature selection step, input features are standardised by removing the mean and scaling to unit variance. The standard score of a sample $x$ is hence calculated as:

\begin{equation}
    z = \frac{x - \hat{\mu}}{\hat{\sigma}}
\end{equation}

where $\hat{\mu}$ and $\hat{\sigma}$ are the mean and the standard deviation of the training samples. This step is performed using the \verb|StandardScaler| implementation provided by the ‘scikit-learn’ Python package \cite{scikit-learn} at the following link: \url{https://github.com/scikit-learn/scikit-learn/blob/98cf537f5/sklearn/preprocessing/_data.py#L644}.

During the stratified \textit{k}-fold cross-validation stage, classes samples are shuffled before splitting into batches. The \verb|random_state| parameter is set to $0$.

Hyper-parameters search is performed using a modified, parallel grid search approach. Also in this case, the basic implementation is provided by the ‘scikit-learn’ Python package \cite{scikit-learn} at the following link: \url{https://github.com/scikit-learn/scikit-learn/blob/98cf537f5/sklearn/model_selection/_search.py#L1031}.

\begin{table}[H]
\centering
\caption{Subset' size-dependent in-sample optimal hyper-parameters configurations and corresponding balanced accuracy scores for $\textrm{Inf-FS}_U$.}
\vspace{0.1in}
\label{tab:Hyperopt_IFS}
\scalebox{0.55}{%
\begin{tabular}{cc|ccc|ccc|ccc|ccc|ccc}
\hline
 &
   &
  \multicolumn{3}{c|}{\textbf{10}} &
  \multicolumn{3}{c|}{\textbf{50}} &
  \multicolumn{3}{c|}{\textbf{100}} &
  \multicolumn{3}{c|}{\textbf{150}} &
  \multicolumn{3}{c}{\textbf{200}} \\ \cline{3-17} 
 &
   &
  \textbf{$\alpha$} &
  \textbf{$\theta$} &
  \textbf{score} &
  \textbf{$\alpha$} &
  \textbf{$\theta$} &
  \textbf{score} &
  \textbf{$\alpha$} &
  \textbf{$\theta$} &
  \textbf{score} &
  \textbf{$\alpha$} &
  \textbf{$\theta$} &
  \textbf{score} &
  \textbf{$\alpha$} &
  \textbf{$\theta$} &
  \textbf{score} \\ \hline
\multicolumn{1}{c|}{\multirow{3}{*}{PCMAC}}       & LinearSVM     & 0.90 & 0.80 & 0.55 & 0.80 & 0.10 & 0.57 & 0.50 & 0.10 & 0.60 & 0.40 & 0.10 & 0.61 & 0.90 & 0.10 & 0.64 \\
\multicolumn{1}{c|}{}                             & KNN           & 0.20 & 0.10 & 0.52 & 0.40 & 0.90 & 0.58 & 0.40 & 0.10 & 0.59 & 0.50 & 0.10 & 0.61 & 0.50 & 0.50 & 0.61 \\
\multicolumn{1}{c|}{}                             & Decision Tree & 0.30 & 0.10 & 0.55 & 0.90 & 0.10 & 0.60 & 0.80 & 0.30 & 0.62 & 0.90 & 0.10 & 0.64 & 0.40 & 0.70 & 0.65 \\ \hline
\multicolumn{1}{c|}{\multirow{3}{*}{RELATHE}}     & LinearSVM     & 0.40 & 0.10 & 0.59 & 0.30 & 0.10 & 0.69 & 0.40 & 0.10 & 0.75 & 0.70 & 0.10 & 0.75 & 0.30 & 0.10 & 0.75 \\
\multicolumn{1}{c|}{}                             & KNN           & 0.80 & 0.10 & 0.59 & 0.30 & 0.30 & 0.67 & 0.50 & 0.80 & 0.72 & 0.40 & 0.10 & 0.72 & 0.40 & 0.10 & 0.73 \\
\multicolumn{1}{c|}{}                             & Decision Tree & 0.40 & 0.10 & 0.58 & 0.40 & 0.90 & 0.67 & 0.50 & 0.30 & 0.72 & 0.60 & 0.30 & 0.74 & 0.40 & 0.20 & 0.74 \\ \hline
\multicolumn{1}{c|}{\multirow{3}{*}{COIL20}}      & LinearSVM     & 0.70 & 0.90 & 0.50 & 0.60 & 0.50 & 0.74 & 0.60 & 0.10 & 0.84 & 0.50 & 0.90 & 0.89 & 0.10 & 0.30 & 0.91 \\
\multicolumn{1}{c|}{}                             & KNN           & 0.70 & 0.10 & 0.67 & 0.70 & 0.10 & 0.85 & 0.50 & 0.10 & 0.89 & 0.50 & 0.10 & 0.92 & 0.10 & 0.10 & 0.94 \\
\multicolumn{1}{c|}{}                             & Decision Tree & 0.70 & 0.10 & 0.66 & 0.70 & 0.30 & 0.82 & 0.40 & 0.80 & 0.83 & 0.20 & 0.80 & 0.85 & 0.10 & 0.30 & 0.86 \\ \hline
\multicolumn{1}{c|}{\multirow{3}{*}{ORL}}         & LinearSVM     & 0.70 & 0.40 & 0.35 & 0.90 & 0.10 & 0.63 & 0.90 & 0.50 & 0.74 & 0.90 & 0.90 & 0.77 & 0.90 & 0.10 & 0.78 \\
\multicolumn{1}{c|}{}                             & KNN           & 0.70 & 0.40 & 0.29 & 0.90 & 0.10 & 0.50 & 0.90 & 0.50 & 0.55 & 0.90 & 0.30 & 0.62 & 0.10 & 0.10 & 0.65 \\
\multicolumn{1}{c|}{}                             & Decision Tree & 0.70 & 0.10 & 0.30 & 0.90 & 0.30 & 0.40 & 0.90 & 0.80 & 0.43 & 0.40 & 0.20 & 0.43 & 0.70 & 0.20 & 0.45 \\ \hline
\multicolumn{1}{c|}{\multirow{3}{*}{warpAR10P}}   & LinearSVM     & 0.70 & 0.10 & 0.39 & 0.10 & 0.40 & 0.61 & 0.40 & 0.50 & 0.71 & 0.60 & 0.60 & 0.78 & 0.20 & 0.30 & 0.82 \\
\multicolumn{1}{c|}{}                             & KNN           & 0.70 & 0.10 & 0.36 & 0.90 & 0.30 & 0.37 & 0.60 & 0.90 & 0.38 & 0.10 & 0.90 & 0.38 & 0.10 & 0.80 & 0.37 \\
\multicolumn{1}{c|}{}                             & Decision Tree & 0.60 & 0.10 & 0.39 & 0.70 & 0.60 & 0.46 & 0.40 & 0.80 & 0.47 & 0.50 & 0.10 & 0.50 & 0.10 & 0.50 & 0.57 \\ \hline
\multicolumn{1}{c|}{\multirow{3}{*}{warpPIE10P}}  & LinearSVM     & 0.10 & 0.30 & 0.78 & 0.70 & 0.10 & 0.97 & 0.80 & 0.10 & 0.99 & 0.80 & 0.10 & 0.99 & 0.10 & 0.10 & 0.99 \\
\multicolumn{1}{c|}{}                             & KNN           & 0.10 & 0.10 & 0.64 & 0.10 & 0.10 & 0.81 & 0.10 & 0.10 & 0.81 & 0.70 & 0.30 & 0.82 & 0.40 & 0.10 & 0.85 \\
\multicolumn{1}{c|}{}                             & Decision Tree & 0.70 & 0.20 & 0.66 & 0.10 & 0.70 & 0.76 & 0.10 & 0.20 & 0.79 & 0.50 & 0.40 & 0.83 & 0.40 & 0.50 & 0.83 \\ \hline
\multicolumn{1}{c|}{\multirow{3}{*}{Yale}}        & LinearSVM     & 0.30 & 0.10 & 0.20 & 0.10 & 0.70 & 0.33 & 0.10 & 0.10 & 0.49 & 0.80 & 0.30 & 0.55 & 0.90 & 0.10 & 0.68 \\
\multicolumn{1}{c|}{}                             & KNN           & 0.70 & 0.10 & 0.25 & 0.50 & 0.10 & 0.34 & 0.10 & 0.10 & 0.36 & 0.30 & 0.10 & 0.42 & 0.40 & 0.20 & 0.49 \\
\multicolumn{1}{c|}{}                             & Decision Tree & 0.90 & 0.10 & 0.31 & 0.60 & 0.40 & 0.39 & 0.30 & 0.40 & 0.40 & 0.90 & 0.20 & 0.46 & 0.80 & 0.80 & 0.48 \\ \hline
\multicolumn{1}{c|}{\multirow{3}{*}{USPS}}        & LinearSVM     & 0.90 & 0.10 & 0.71 & 0.90 & 0.10 & 0.89 & 0.90 & 0.10 & 0.91 & 0.70 & 0.40 & 0.92 & 0.10 & 0.10 & 0.92 \\
\multicolumn{1}{c|}{}                             & KNN           & 0.90 & 0.10 & 0.76 & 0.90 & 0.10 & 0.93 & 0.90 & 0.10 & 0.96 & 0.80 & 0.10 & 0.96 & 0.10 & 0.10 & 0.95 \\
\multicolumn{1}{c|}{}                             & Decision Tree & 0.90 & 0.10 & 0.70 & 0.90 & 0.10 & 0.83 & 0.90 & 0.10 & 0.85 & 0.40 & 0.10 & 0.86 & 0.20 & 0.80 & 0.86 \\ \hline
\multicolumn{1}{c|}{\multirow{3}{*}{colon}}       & LinearSVM     & 0.60 & 0.10 & 0.72 & 0.40 & 0.50 & 0.81 & 0.60 & 0.90 & 0.78 & 0.80 & 0.50 & 0.76 & 0.90 & 0.40 & 0.81 \\
\multicolumn{1}{c|}{}                             & KNN           & 0.70 & 0.10 & 0.74 & 0.80 & 0.40 & 0.80 & 0.80 & 0.10 & 0.80 & 0.80 & 0.10 & 0.86 & 0.70 & 0.10 & 0.83 \\
\multicolumn{1}{c|}{}                             & Decision Tree & 0.60 & 0.10 & 0.72 & 0.80 & 0.90 & 0.81 & 0.80 & 0.40 & 0.83 & 0.30 & 0.70 & 0.79 & 0.20 & 0.90 & 0.83 \\ \hline
\multicolumn{1}{c|}{\multirow{3}{*}{GLIOMA}}      & LinearSVM     & 0.90 & 0.10 & 0.55 & 0.10 & 0.20 & 0.62 & 0.60 & 0.10 & 0.63 & 0.60 & 0.20 & 0.70 & 0.70 & 0.10 & 0.67 \\
\multicolumn{1}{c|}{}                             & KNN           & 0.90 & 0.10 & 0.50 & 0.90 & 0.10 & 0.68 & 0.90 & 0.10 & 0.70 & 0.90 & 0.60 & 0.69 & 0.90 & 0.10 & 0.69 \\
\multicolumn{1}{c|}{}                             & Decision Tree & 0.80 & 0.50 & 0.53 & 0.20 & 0.10 & 0.68 & 0.50 & 0.10 & 0.59 & 0.30 & 0.80 & 0.58 & 0.40 & 0.40 & 0.58 \\ \hline
\multicolumn{1}{c|}{\multirow{3}{*}{lung}}        & LinearSVM     & 0.90 & 0.10 & 0.45 & 0.90 & 0.10 & 0.78 & 0.80 & 0.10 & 0.90 & 0.80 & 0.10 & 0.93 & 0.60 & 0.10 & 0.88 \\
\multicolumn{1}{c|}{}                             & KNN           & 0.90 & 0.10 & 0.36 & 0.80 & 0.10 & 0.70 & 0.90 & 0.10 & 0.75 & 0.80 & 0.10 & 0.81 & 0.70 & 0.10 & 0.75 \\
\multicolumn{1}{c|}{}                             & Decision Tree & 0.90 & 0.10 & 0.32 & 0.80 & 0.10 & 0.56 & 0.60 & 0.30 & 0.69 & 0.60 & 0.80 & 0.71 & 0.90 & 0.10 & 0.76 \\ \hline
\multicolumn{1}{c|}{\multirow{3}{*}{lung\_small}} & LinearSVM     & 0.70 & 0.10 & 0.62 & 0.10 & 0.10 & 0.72 & 0.10 & 0.90 & 0.71 & 0.90 & 0.80 & 0.76 & 0.10 & 0.40 & 0.75 \\
\multicolumn{1}{c|}{}                             & KNN           & 0.70 & 0.10 & 0.67 & 0.30 & 0.50 & 0.80 & 0.30 & 0.10 & 0.83 & 0.40 & 0.50 & 0.83 & 0.10 & 0.10 & 0.74 \\
\multicolumn{1}{c|}{}                             & Decision Tree & 0.10 & 0.80 & 0.55 & 0.10 & 0.40 & 0.60 & 0.30 & 0.90 & 0.61 & 0.30 & 0.30 & 0.64 & 0.20 & 0.10 & 0.69 \\ \hline
\multicolumn{1}{c|}{\multirow{3}{*}{lymphoma}}    & LinearSVM     & 0.90 & 0.10 & 0.38 & 0.60 & 0.10 & 0.52 & 0.70 & 0.10 & 0.65 & 0.60 & 0.10 & 0.66 & 0.30 & 0.90 & 0.71 \\
\multicolumn{1}{c|}{}                             & KNN           & 0.90 & 0.90 & 0.27 & 0.90 & 0.10 & 0.44 & 0.80 & 0.50 & 0.61 & 0.70 & 0.10 & 0.62 & 0.80 & 0.10 & 0.66 \\
\multicolumn{1}{c|}{}                             & Decision Tree & 0.20 & 0.60 & 0.33 & 0.80 & 0.70 & 0.51 & 0.70 & 0.70 & 0.50 & 0.60 & 0.20 & 0.55 & 0.60 & 0.50 & 0.60 \\ \hline
\multicolumn{1}{c|}{\multirow{3}{*}{GISETTE}}     & LinearSVM     & 0.90 & 0.10 & 0.85 & 0.90 & 0.10 & 0.91 & 0.80 & 0.10 & 0.92 & 0.80 & 0.10 & 0.93 & 0.70 & 0.10 & 0.93 \\
\multicolumn{1}{c|}{}                             & KNN           & 0.90 & 0.10 & 0.86 & 0.90 & 0.10 & 0.94 & 0.90 & 0.10 & 0.94 & 0.80 & 0.10 & 0.94 & 0.80 & 0.10 & 0.94 \\
\multicolumn{1}{c|}{}                             & Decision Tree & 0.90 & 0.10 & 0.80 & 0.80 & 0.50 & 0.88 & 0.80 & 0.50 & 0.89 & 0.90 & 0.60 & 0.90 & 0.80 & 0.10 & 0.90 \\ \hline
\multicolumn{1}{c|}{\multirow{3}{*}{Isolet}}      & LinearSVM     & 0.60 & 0.10 & 0.31 & 0.80 & 0.10 & 0.72 & 0.90 & 0.10 & 0.79 & 0.90 & 0.30 & 0.86 & 0.90 & 0.10 & 0.86 \\
\multicolumn{1}{c|}{}                             & KNN           & 0.60 & 0.10 & 0.27 & 0.90 & 0.10 & 0.70 & 0.90 & 0.50 & 0.77 & 0.90 & 0.10 & 0.80 & 0.80 & 0.30 & 0.81 \\
\multicolumn{1}{c|}{}                             & Decision Tree & 0.80 & 0.10 & 0.24 & 0.90 & 0.40 & 0.68 & 0.80 & 0.50 & 0.71 & 0.90 & 0.30 & 0.75 & 0.90 & 0.70 & 0.75 \\ \hline
\multicolumn{1}{c|}{\multirow{3}{*}{MADELON}}     & LinearSVM     & 0.60 & 0.10 & 0.61 & 0.80 & 0.10 & 0.60 & 0.70 & 0.10 & 0.59 & 0.30 & 0.10 & 0.57 & 0.30 & 0.10 & 0.55 \\
\multicolumn{1}{c|}{}                             & KNN           & 0.80 & 0.10 & 0.71 & 0.80 & 0.10 & 0.61 & 0.90 & 0.10 & 0.59 & 0.90 & 0.10 & 0.57 & 0.80 & 0.10 & 0.57 \\
\multicolumn{1}{c|}{}                             & Decision Tree & 0.80 & 0.10 & 0.70 & 0.80 & 0.10 & 0.70 & 0.80 & 0.80 & 0.75 & 0.70 & 0.70 & 0.75 & 0.60 & 0.60 & 0.74 \\ \hline
\end{tabular}%
}
\end{table}

\begin{table}[H]
\caption{Subset' size-dependent in-sample optimal hyper-parameters configurations and corresponding balanced accuracy scores for TFS.}
\label{tab:Hyperopt_TFS}
\vspace{0.1in}
\centering
\scalebox{0.5}{%
\begin{tabular}{cc|cccc|cccc|cccc|cccc|cccc}
\hline
 &
   &
  \multicolumn{4}{c|}{\textbf{10}} &
  \multicolumn{4}{c|}{\textbf{50}} &
  \multicolumn{4}{c|}{\textbf{100}} &
  \multicolumn{4}{c|}{\textbf{150}} &
  \multicolumn{4}{c}{\textbf{200}} \\ \cline{3-22} 
 &
   &
  \textbf{metric} &
  \textbf{square} &
  \textbf{$\alpha$} &
  \textbf{score} &
  \textbf{metric} &
  \textbf{square} &
  \textbf{$\alpha$} &
  \textbf{score} &
  \textbf{metric} &
  \textbf{square} &
  \textbf{$\alpha$} &
  \textbf{score} &
  \textbf{metric} &
  \textbf{square} &
  \textbf{$\alpha$} &
  \textbf{score} &
  \textbf{metric} &
  \textbf{square} &
  \textbf{$\alpha$} &
  \textbf{score} \\ \hline
\multicolumn{1}{c|}{\multirow{3}{*}{PCMAC}} &
  LinearSVM &
  Spearman &
  Square &
  / &
  0.56 &
  Pearson &
  Normal &
  / &
  0.62 &
  Pearson &
  Normal &
  / &
  0.64 &
  Pearson &
  Square &
  / &
  0.66 &
  Energy &
  Normal &
  0.30 &
  0.69 \\
\multicolumn{1}{c|}{} &
  KNN &
  Pearson &
  Normal &
  / &
  0.52 &
  Pearson &
  Normal &
  / &
  0.60 &
  Pearson &
  Normal &
  / &
  0.59 &
  Pearson &
  Square &
  / &
  0.62 &
  Energy &
  Normal &
  0.40 &
  0.64 \\
\multicolumn{1}{c|}{} &
  Decision Tree &
  Spearman &
  Square &
  / &
  0.56 &
  Pearson &
  Normal &
  / &
  0.63 &
  Pearson &
  Square &
  / &
  0.64 &
  Pearson &
  Square &
  / &
  0.66 &
  Pearson &
  Normal &
  / &
  0.68 \\ \hline
\multicolumn{1}{c|}{\multirow{3}{*}{RELATHE}} &
  LinearSVM &
  Spearman &
  Square &
  / &
  0.53 &
  Energy &
  Normal &
  0.70 &
  0.62 &
  Energy &
  Normal &
  0.70 &
  0.69 &
  Energy &
  Normal &
  0.90 &
  0.70 &
  Energy &
  Normal &
  0.90 &
  0.71 \\
\multicolumn{1}{c|}{} &
  KNN &
  Energy &
  Normal &
  0.30 &
  0.53 &
  Pearson &
  Normal &
  / &
  0.61 &
  Energy &
  Normal &
  0.80 &
  0.64 &
  Energy &
  Normal &
  0.30 &
  0.66 &
  Energy &
  Normal &
  0.60 &
  0.68 \\
\multicolumn{1}{c|}{} &
  Decision Tree &
  Energy &
  Normal &
  0.30 &
  0.53 &
  Energy &
  Normal &
  0.60 &
  0.60 &
  Spearman &
  Normal &
  / &
  0.64 &
  Energy &
  Normal &
  0.80 &
  0.68 &
  Spearman &
  Square &
  / &
  0.67 \\ \hline
\multicolumn{1}{c|}{\multirow{3}{*}{COIL20}} &
  LinearSVM &
  Pearson &
  Square &
  / &
  0.71 &
  Pearson &
  Square &
  / &
  0.91 &
  Energy &
  Normal &
  0.20 &
  0.93 &
  Pearson &
  Square &
  / &
  0.95 &
  Pearson &
  Normal &
  / &
  0.95 \\
\multicolumn{1}{c|}{} &
  KNN &
  Pearson &
  Square &
  / &
  0.81 &
  Pearson &
  Square &
  / &
  0.92 &
  Spearman &
  Normal &
  / &
  0.92 &
  Spearman &
  Normal &
  / &
  0.93 &
  Pearson &
  Normal &
  / &
  0.92 \\
\multicolumn{1}{c|}{} &
  Decision Tree &
  Pearson &
  Square &
  / &
  0.81 &
  Pearson &
  Square &
  / &
  0.89 &
  Pearson &
  Normal &
  / &
  0.89 &
  Pearson &
  Square &
  / &
  0.89 &
  Pearson &
  Normal &
  / &
  0.89 \\ \hline
\multicolumn{1}{c|}{\multirow{3}{*}{ORL}} &
  LinearSVM &
  Pearson &
  Normal &
  / &
  0.50 &
  Spearman &
  Normal &
  / &
  0.84 &
  Spearman &
  Normal &
  / &
  0.89 &
  Spearman &
  Normal &
  / &
  0.88 &
  Spearman &
  Normal &
  / &
  0.89 \\
\multicolumn{1}{c|}{} &
  KNN &
  Spearman &
  Square &
  / &
  0.43 &
  Spearman &
  Square &
  / &
  0.57 &
  Energy &
  Normal &
  0.50 &
  0.64 &
  Energy &
  Normal &
  0.20 &
  0.66 &
  Energy &
  Normal &
  0.30 &
  0.68 \\
\multicolumn{1}{c|}{} &
  Decision Tree &
  Pearson &
  Normal &
  / &
  0.40 &
  Spearman &
  Square &
  / &
  0.45 &
  Pearson &
  Normal &
  / &
  0.48 &
  Pearson &
  Normal &
  / &
  0.51 &
  Pearson &
  Normal &
  / &
  0.53 \\ \hline
\multicolumn{1}{c|}{\multirow{3}{*}{warpAR10P}} &
  LinearSVM &
  Energy &
  Normal &
  0.60 &
  0.46 &
  Energy &
  Normal &
  0.40 &
  0.74 &
  Energy &
  Normal &
  0.70 &
  0.85 &
  Pearson &
  Square &
  / &
  0.86 &
  Energy &
  Normal &
  0.60 &
  0.89 \\
\multicolumn{1}{c|}{} &
  KNN &
  Energy &
  Normal &
  0.70 &
  0.38 &
  Energy &
  Normal &
  0.10 &
  0.42 &
  Energy &
  Normal &
  0.10 &
  0.45 &
  Energy &
  Normal &
  0.20 &
  0.49 &
  Energy &
  Normal &
  0.10 &
  0.47 \\
\multicolumn{1}{c|}{} &
  Decision Tree &
  Energy &
  Normal &
  0.50 &
  0.46 &
  Energy &
  Normal &
  0.10 &
  0.70 &
  Energy &
  Normal &
  0.10 &
  0.61 &
  Energy &
  Normal &
  0.10 &
  0.64 &
  Energy &
  Normal &
  0.60 &
  0.65 \\ \hline
\multicolumn{1}{c|}{\multirow{3}{*}{warpPIE10P}} &
  LinearSVM &
  Energy &
  Normal &
  0.20 &
  0.78 &
  Energy &
  Normal &
  0.20 &
  0.98 &
  Energy &
  Normal &
  0.10 &
  0.99 &
  Energy &
  Normal &
  0.20 &
  1.00 &
  Energy &
  Normal &
  0.10 &
  1.00 \\
\multicolumn{1}{c|}{} &
  KNN &
  Energy &
  Normal &
  0.20 &
  0.71 &
  Energy &
  Normal &
  0.40 &
  0.86 &
  Energy &
  Normal &
  0.10 &
  0.86 &
  Energy &
  Normal &
  0.20 &
  0.87 &
  Energy &
  Normal &
  0.10 &
  0.87 \\
\multicolumn{1}{c|}{} &
  Decision Tree &
  Energy &
  Normal &
  0.10 &
  0.66 &
  Energy &
  Normal &
  0.30 &
  0.75 &
  Energy &
  Normal &
  0.40 &
  0.81 &
  Energy &
  Normal &
  0.70 &
  0.84 &
  Energy &
  Normal &
  0.10 &
  0.80 \\ \hline
\multicolumn{1}{c|}{\multirow{3}{*}{Yale}} &
  LinearSVM &
  Spearman &
  Normal &
  / &
  0.49 &
  Pearson &
  Normal &
  / &
  0.63 &
  Pearson &
  Normal &
  / &
  0.69 &
  Pearson &
  Square &
  / &
  0.73 &
  Energy &
  Normal &
  0.20 &
  0.74 \\
\multicolumn{1}{c|}{} &
  KNN &
  Spearman &
  Normal &
  / &
  0.43 &
  Pearson &
  Square &
  / &
  0.56 &
  Spearman &
  Normal &
  / &
  0.56 &
  Pearson &
  Normal &
  / &
  0.57 &
  Spearman &
  Square &
  / &
  0.60 \\
\multicolumn{1}{c|}{} &
  Decision Tree &
  Pearson &
  Square &
  / &
  0.38 &
  Pearson &
  Square &
  / &
  0.49 &
  Pearson &
  Normal &
  / &
  0.48 &
  Spearman &
  Normal &
  / &
  0.56 &
  Pearson &
  Normal &
  / &
  0.52 \\ \hline
\multicolumn{1}{c|}{\multirow{3}{*}{USPS}} &
  LinearSVM &
  Energy &
  Normal &
  0.70 &
  0.71 &
  Energy &
  Normal &
  0.20 &
  0.90 &
  Energy &
  Normal &
  0.10 &
  0.92 &
  Pearson &
  Normal &
  / &
  0.92 &
  Pearson &
  Normal &
  / &
  0.92 \\
\multicolumn{1}{c|}{} &
  KNN &
  Energy &
  Normal &
  0.50 &
  0.77 &
  Energy &
  Normal &
  0.90 &
  0.94 &
  Energy &
  Normal &
  0.10 &
  0.94 &
  Energy &
  Normal &
  0.30 &
  0.95 &
  Spearman &
  Normal &
  / &
  0.95 \\
\multicolumn{1}{c|}{} &
  Decision Tree &
  Spearman &
  Square &
  / &
  0.71 &
  Energy &
  Normal &
  0.90 &
  0.84 &
  Energy &
  Normal &
  0.70 &
  0.85 &
  Pearson &
  Normal &
  / &
  0.85 &
  Energy &
  Normal &
  0.30 &
  0.86 \\ \hline
\multicolumn{1}{c|}{\multirow{3}{*}{colon}} &
  LinearSVM &
  Energy &
  Normal &
  0.30 &
  0.73 &
  Energy &
  Normal &
  0.80 &
  0.71 &
  Energy &
  Normal &
  0.90 &
  0.74 &
  Energy &
  Normal &
  0.90 &
  0.79 &
  Energy &
  Normal &
  0.60 &
  0.77 \\
\multicolumn{1}{c|}{} &
  KNN &
  Energy &
  Normal &
  0.90 &
  0.83 &
  Energy &
  Normal &
  0.30 &
  0.76 &
  Energy &
  Normal &
  0.30 &
  0.80 &
  Energy &
  Normal &
  0.90 &
  0.76 &
  Energy &
  Normal &
  0.90 &
  0.82 \\
\multicolumn{1}{c|}{} &
  Decision Tree &
  Pearson &
  Normal &
  / &
  0.67 &
  Energy &
  Normal &
  0.90 &
  0.70 &
  Energy &
  Normal &
  0.60 &
  0.76 &
  Energy &
  Normal &
  0.50 &
  0.71 &
  Energy &
  Normal &
  0.50 &
  0.76 \\ \hline
\multicolumn{1}{c|}{\multirow{3}{*}{GLIOMA}} &
  LinearSVM &
  Pearson &
  Square &
  / &
  0.65 &
  Pearson &
  Normal &
  / &
  0.69 &
  Energy &
  Normal &
  0.20 &
  0.71 &
  Energy &
  Normal &
  0.50 &
  0.76 &
  Pearson &
  Square &
  / &
  0.72 \\
\multicolumn{1}{c|}{} &
  KNN &
  Pearson &
  Normal &
  / &
  0.58 &
  Energy &
  Normal &
  0.80 &
  0.72 &
  Pearson &
  Normal &
  / &
  0.68 &
  Pearson &
  Normal &
  / &
  0.77 &
  Pearson &
  Normal &
  / &
  0.71 \\
\multicolumn{1}{c|}{} &
  Decision Tree &
  Pearson &
  Square &
  / &
  0.60 &
  Energy &
  Normal &
  0.70 &
  0.72 &
  Energy &
  Normal &
  0.40 &
  0.75 &
  Energy &
  Normal &
  0.40 &
  0.69 &
  Energy &
  Normal &
  0.70 &
  0.62 \\ \hline
\multicolumn{1}{c|}{\multirow{3}{*}{lung}} &
  LinearSVM &
  Spearman &
  Square &
  / &
  0.74 &
  Pearson &
  Normal &
  / &
  0.96 &
  Pearson &
  Normal &
  / &
  0.97 &
  Spearman &
  Normal &
  / &
  0.97 &
  Energy &
  Normal &
  0.90 &
  0.94 \\
\multicolumn{1}{c|}{} &
  KNN &
  Spearman &
  Square &
  / &
  0.61 &
  Pearson &
  Square &
  / &
  0.79 &
  Energy &
  Normal &
  0.40 &
  0.76 &
  Energy &
  Normal &
  0.60 &
  0.79 &
  Spearman &
  Square &
  / &
  0.80 \\
\multicolumn{1}{c|}{} &
  Decision Tree &
  Pearson &
  Square &
  / &
  0.75 &
  Spearman &
  Normal &
  / &
  0.87 &
  Pearson &
  Normal &
  / &
  0.87 &
  Pearson &
  Normal &
  / &
  0.79 &
  Pearson &
  Normal &
  / &
  0.90 \\ \hline
\multicolumn{1}{c|}{\multirow{3}{*}{lung\_small}} &
  LinearSVM &
  Energy &
  Normal &
  0.80 &
  0.54 &
  Energy &
  Normal &
  0.80 &
  0.72 &
  Pearson &
  Square &
  / &
  0.76 &
  Pearson &
  Normal &
  / &
  0.75 &
  Spearman &
  Square &
  / &
  0.79 \\
\multicolumn{1}{c|}{} &
  KNN &
  Energy &
  Normal &
  0.60 &
  0.59 &
  Energy &
  Normal &
  0.60 &
  0.77 &
  Energy &
  Normal &
  0.10 &
  0.79 &
  Energy &
  Normal &
  0.40 &
  0.76 &
  Energy &
  Normal &
  0.20 &
  0.71 \\
\multicolumn{1}{c|}{} &
  Decision Tree &
  Spearman &
  Normal &
  / &
  0.48 &
  Energy &
  Normal &
  0.80 &
  0.57 &
  Pearson &
  Normal &
  / &
  0.58 &
  Energy &
  Normal &
  0.80 &
  0.56 &
  Energy &
  Normal &
  0.10 &
  0.68 \\ \hline
\multicolumn{1}{c|}{\multirow{3}{*}{lymphoma}} &
  LinearSVM &
  Pearson &
  Square &
  / &
  0.47 &
  Energy &
  Normal &
  0.70 &
  0.58 &
  Energy &
  Normal &
  0.60 &
  0.66 &
  Energy &
  Normal &
  0.20 &
  0.75 &
  Energy &
  Normal &
  0.50 &
  0.79 \\
\multicolumn{1}{c|}{} &
  KNN &
  Pearson &
  Square &
  / &
  0.41 &
  Spearman &
  Square &
  / &
  0.54 &
  Energy &
  Normal &
  0.70 &
  0.61 &
  Energy &
  Normal &
  0.70 &
  0.67 &
  Energy &
  Normal &
  0.70 &
  0.63 \\
\multicolumn{1}{c|}{} &
  Decision Tree &
  Pearson &
  Square &
  / &
  0.44 &
  Energy &
  Normal &
  0.20 &
  0.57 &
  Energy &
  Normal &
  0.30 &
  0.50 &
  Pearson &
  Normal &
  / &
  0.57 &
  Energy &
  Normal &
  0.20 &
  0.56 \\ \hline
\multicolumn{1}{c|}{\multirow{3}{*}{GISETTE}} &
  LinearSVM &
  Energy &
  Normal &
  0.20 &
  0.66 &
  Energy &
  Normal &
  0.40 &
  0.81 &
  Energy &
  Normal &
  0.40 &
  0.87 &
  Energy &
  Normal &
  0.40 &
  0.90 &
  Energy &
  Normal &
  0.40 &
  0.91 \\
\multicolumn{1}{c|}{} &
  KNN &
  Energy &
  Normal &
  0.20 &
  0.59 &
  Energy &
  Normal &
  0.90 &
  0.77 &
  Energy &
  Normal &
  0.40 &
  0.80 &
  Energy &
  Normal &
  0.40 &
  0.84 &
  Energy &
  Normal &
  0.40 &
  0.84 \\
\multicolumn{1}{c|}{} &
  Decision Tree &
  Energy &
  Normal &
  0.20 &
  0.61 &
  Energy &
  Normal &
  0.90 &
  0.79 &
  Energy &
  Normal &
  0.40 &
  0.85 &
  Energy &
  Normal &
  0.40 &
  0.89 &
  Energy &
  Normal &
  0.40 &
  0.89 \\ \hline
\multicolumn{1}{c|}{\multirow{3}{*}{Isolet}} &
  LinearSVM &
  Spearman &
  Square &
  / &
  0.51 &
  Pearson &
  Normal &
  / &
  0.74 &
  Pearson &
  Normal &
  / &
  0.81 &
  Pearson &
  Square &
  / &
  0.88 &
  Pearson &
  Normal &
  / &
  0.89 \\
\multicolumn{1}{c|}{} &
  KNN &
  Pearson &
  Square &
  / &
  0.51 &
  Pearson &
  Normal &
  / &
  0.69 &
  Spearman &
  Normal &
  / &
  0.74 &
  Spearman &
  Normal &
  / &
  0.79 &
  Spearman &
  Normal &
  / &
  0.81 \\
\multicolumn{1}{c|}{} &
  Decision Tree &
  Pearson &
  Normal &
  / &
  0.46 &
  Pearson &
  Normal &
  / &
  0.65 &
  Pearson &
  Normal &
  / &
  0.70 &
  Pearson &
  Normal &
  / &
  0.73 &
  Pearson &
  Square &
  / &
  0.75 \\ \cline{1-8} \cline{10-22} 
\multicolumn{1}{c|}{\multirow{3}{*}{MADELON}} &
  LinearSVM &
  Pearson &
  Square &
  / &
  0.61 &
  Energy &
  Normal &
  0.30 &
  0.60 &
  Pearson &
  Square &
  / &
  0.58 &
  Spearman &
  Square &
  / &
  0.57 &
  Energy &
  Normal &
  0.30 &
  0.57 \\
\multicolumn{1}{c|}{} &
  KNN &
  Spearman &
  Normal &
  / &
  0.75 &
  Pearson &
  Square &
  / &
  0.72 &
  Spearman &
  Square &
  / &
  0.66 &
  Spearman &
  Square &
  / &
  0.62 &
  Pearson &
  Normal &
  / &
  0.59 \\
\multicolumn{1}{c|}{} &
  Decision Tree &
  Pearson &
  Normal &
  / &
  0.72 &
  Spearman &
  Normal &
  / &
  0.79 &
  Spearman &
  Normal &
  / &
  0.76 &
  Spearman &
  Normal &
  / &
  0.77 &
  Spearman &
  Normal &
  / &
  0.75 \\ \hline
\end{tabular}%
}
\end{table}

\newpage

\section{Additional out-of-sample evaluations} \label{Appendix_6}
To prevent from obtaining metric-dependent out-of-sample results, in addition to the Balanced Accuracy score (see Equation \ref{eq:BA_multi}), we consider two additional metrics: (i) the F1 score (ii) and the Matthews Correlation Coefficient \cite{matthews1975comparison}. \\

The general formulation for the F1 score is

\begin{equation} \label{eq:F1_multi}
    F1 = 2 \times \frac{ \frac{1}{|Z|} \sum_{z \in Z}(\frac{\textrm{TP}_z}{\textrm{TP}_z + \textrm{FP}_z}) \times \frac{1}{|Z|} \sum_{z \in Z}(\frac{\textrm{TP}_z}{\textrm{TP}_z + \textrm{FN}_z})}{\frac{1}{|Z|} \sum_{z \in Z}(\frac{\textrm{TP}_z}{\textrm{TP}_z + \textrm{FP}_z}) + \frac{1}{|Z|} \sum_{z \in Z}(\frac{\textrm{TP}_z}{\textrm{TP}_z + \textrm{FN}_z})}.
\end{equation}

$\textrm{TP}$ is the number of outcomes where the model correctly classifies a sample as belonging to a positive class (or detects an event of interest), when in fact it does belong to that class (or the event is present). $\textrm{FP}$ is the number of outcomes where the model incorrectly classifies a sample as belonging to a positive class (or detects an event of interest), when in fact it does not belong to that class (or the event is not present). $\textrm{FN}$ is the number of outcomes where the model incorrectly classifies a sample as belonging to a negative class (or fails to detect an event of interest), when in fact it belongs to a positive class (or the event of interest is present). $|Z|$ indicates the cardinality of the set of different classes.

The general formulation for the MCC is

\begin{equation} \label{eq:MCC_multi}
    MCC = \frac{(C \times S) - (T \cdot P)}{\sqrt{S^2 - (P \cdot P)} \times \sqrt{S^2 - (T \cdot T)}}
\end{equation}

where $T$ is a vector containing the number of times each class $z \in Z$ truly occurs, $P$ is a vector containing the number of times each class $z \in Z$ is predicted, $C$ is the total number of samples correctly predicted and $S$ is the total number of samples. Given Equations \ref{eq:F1_multi} and \ref{eq:MCC_multi}, it is easy for the interested reader to reconstruct the formulation for the binary case.

For each performance metric, we use the implementation provided by the ‘scikit-learn’ Python package \cite{scikit-learn} at the following links:

\begin{itemize}
    \item Balanced Accuracy score: \url{https://github.com/scikit-learn/scikit-learn/blob/98cf537f5/sklearn/metrics/_classification.py#L2111}
    \item F1 score: \url{https://github.com/scikit-learn/scikit-learn/blob/98cf537f5/sklearn/metrics/_classification.py#L1011}
    \item Matthews Correlation Coefficient: \url{https://github.com/scikit-learn/scikit-learn/blob/98cf537f5/sklearn/metrics/_classification.py#L848}
\end{itemize}

\begin{table}[H]
\centering
\caption{Subset size-dependent, out-of-sample results using a LinearSVM classifier. We use three evaluation metrics: balanced accuracy score, F1 score and Matthews Correlation Coefficient. For each dataset, we boldly highlight the combination between feature selection schema and classifier producing the best out-of-sample result.}
\label{tab:local_bests_all_metrics_LinearSVM}
\vspace{0.1in}
\scalebox{0.75}{
\begin{tabular}{cc|cccccccccc}
\hline
 &
   &
  \multicolumn{10}{c}{\textbf{Linear SVM}} \\ \cline{3-12} 
 &
   &
  \multicolumn{2}{c|}{\textbf{10}} &
  \multicolumn{2}{c|}{\textbf{50}} &
  \multicolumn{2}{c|}{\textbf{100}} &
  \multicolumn{2}{c|}{\textbf{150}} &
  \multicolumn{2}{c}{\textbf{200}} \\ \cline{3-12} 
 &
   &
  \textbf{$\textrm{Inf-FS}_U$} &
  \multicolumn{1}{c|}{\textbf{TFS}} &
  \textbf{$\textrm{Inf-FS}_U$} &
  \multicolumn{1}{c|}{\textbf{TFS}} &
  \textbf{$\textrm{Inf-FS}_U$} &
  \multicolumn{1}{c|}{\textbf{TFS}} &
  \textbf{$\textrm{Inf-FS}_U$} &
  \multicolumn{1}{c|}{\textbf{TFS}} &
  \textbf{$\textrm{Inf-FS}_U$} &
  \textbf{TFS} \\ \hline
\multicolumn{1}{c|}{\multirow{3}{*}{PCMAC}} &
  Balanced Accuracy &
  0.52 &
  \multicolumn{1}{c|}{0.50} &
  0.57 &
  \multicolumn{1}{c|}{0.67} &
  0.59 &
  \multicolumn{1}{c|}{0.70} &
  0.61 &
  \multicolumn{1}{c|}{\textbf{0.71}} &
  0.62 &
  0.69 \\
\multicolumn{1}{c|}{} &
  F1 Score &
  0.52 &
  \multicolumn{1}{c|}{0.35} &
  0.57 &
  \multicolumn{1}{c|}{0.67} &
  0.59 &
  \multicolumn{1}{c|}{0.70} &
  0.61 &
  \multicolumn{1}{c|}{\textbf{0.71}} &
  0.62 &
  0.69 \\
\multicolumn{1}{c|}{} &
  MCC &
  0.05 &
  \multicolumn{1}{c|}{-0.02} &
  0.13 &
  \multicolumn{1}{c|}{0.34} &
  0.19 &
  \multicolumn{1}{c|}{0.41} &
  0.23 &
  \multicolumn{1}{c|}{\textbf{0.44}} &
  0.25 &
  0.39 \\ \hline
\multicolumn{1}{c|}{\multirow{3}{*}{RELATHE}} &
  Balanced Accuracy &
  0.47 &
  \multicolumn{1}{c|}{0.49} &
  0.43 &
  \multicolumn{1}{c|}{\textbf{0.53}} &
  0.51 &
  \multicolumn{1}{c|}{0.53} &
  0.44 &
  \multicolumn{1}{c|}{0.49} &
  0.53 &
  0.53 \\
\multicolumn{1}{c|}{} &
  F1 Score &
  0.33 &
  \multicolumn{1}{c|}{0.37} &
  0.40 &
  \multicolumn{1}{c|}{0.48} &
  0.50 &
  \multicolumn{1}{c|}{0.51} &
  0.43 &
  \multicolumn{1}{c|}{0.48} &
  0.50 &
  \textbf{0.53} \\
\multicolumn{1}{c|}{} &
  MCC &
  -0.14 &
  \multicolumn{1}{c|}{-0.06} &
  -0.15 &
  \multicolumn{1}{c|}{0.06} &
  0.01 &
  \multicolumn{1}{c|}{\textbf{0.07}} &
  -0.12 &
  \multicolumn{1}{c|}{-0.01} &
  0.07 &
  0.07 \\ \hline
\multicolumn{1}{c|}{\multirow{3}{*}{COIL20}} &
  Balanced Accuracy &
  0.52 &
  \multicolumn{1}{c|}{0.63} &
  0.77 &
  \multicolumn{1}{c|}{0.90} &
  0.84 &
  \multicolumn{1}{c|}{0.92} &
  0.90 &
  \multicolumn{1}{c|}{0.94} &
  0.94 &
  \textbf{0.96} \\
\multicolumn{1}{c|}{} &
  F1 Score &
  0.44 &
  \multicolumn{1}{c|}{0.58} &
  0.76 &
  \multicolumn{1}{c|}{0.90} &
  0.83 &
  \multicolumn{1}{c|}{0.92} &
  0.89 &
  \multicolumn{1}{c|}{0.94} &
  0.94 &
  \textbf{0.95} \\
\multicolumn{1}{c|}{} &
  MCC &
  0.50 &
  \multicolumn{1}{c|}{0.62} &
  0.76 &
  \multicolumn{1}{c|}{0.90} &
  0.83 &
  \multicolumn{1}{c|}{0.92} &
  0.89 &
  \multicolumn{1}{c|}{0.94} &
  0.94 &
  \textbf{0.95} \\ \hline
\multicolumn{1}{c|}{\multirow{3}{*}{ORL}} &
  Balanced Accuracy &
  0.40 &
  \multicolumn{1}{c|}{0.44} &
  0.63 &
  \multicolumn{1}{c|}{0.88} &
  0.72 &
  \multicolumn{1}{c|}{0.89} &
  0.86 &
  \multicolumn{1}{c|}{0.93} &
  0.84 &
  \textbf{0.94} \\
\multicolumn{1}{c|}{} &
  F1 Score &
  0.33 &
  \multicolumn{1}{c|}{0.39} &
  0.61 &
  \multicolumn{1}{c|}{0.86} &
  0.71 &
  \multicolumn{1}{c|}{0.88} &
  0.85 &
  \multicolumn{1}{c|}{\textbf{0.93}} &
  0.84 &
  0.93 \\
\multicolumn{1}{c|}{} &
  MCC &
  0.39 &
  \multicolumn{1}{c|}{0.43} &
  0.63 &
  \multicolumn{1}{c|}{0.87} &
  0.71 &
  \multicolumn{1}{c|}{0.89} &
  0.86 &
  \multicolumn{1}{c|}{0.93} &
  0.84 &
  \textbf{0.94} \\ \hline
\multicolumn{1}{c|}{\multirow{3}{*}{warpAR10P}} &
  Balanced Accuracy &
  0.33 &
  \multicolumn{1}{c|}{0.44} &
  0.56 &
  \multicolumn{1}{c|}{0.78} &
  0.72 &
  \multicolumn{1}{c|}{0.85} &
  0.70 &
  \multicolumn{1}{c|}{\textbf{0.95}} &
  0.75 &
  0.85 \\
\multicolumn{1}{c|}{} &
  F1 Score &
  0.29 &
  \multicolumn{1}{c|}{0.43} &
  0.56 &
  \multicolumn{1}{c|}{0.76} &
  0.71 &
  \multicolumn{1}{c|}{0.85} &
  0.69 &
  \multicolumn{1}{c|}{\textbf{0.94}} &
  0.74 &
  0.84 \\
\multicolumn{1}{c|}{} &
  MCC &
  0.27 &
  \multicolumn{1}{c|}{0.38} &
  0.52 &
  \multicolumn{1}{c|}{0.75} &
  0.69 &
  \multicolumn{1}{c|}{0.83} &
  0.67 &
  \multicolumn{1}{c|}{\textbf{0.95}} &
  0.72 &
  0.84 \\ \hline
\multicolumn{1}{c|}{\multirow{3}{*}{warpPIE10P}} &
  Balanced Accuracy &
  0.85 &
  \multicolumn{1}{c|}{0.89} &
  0.95 &
  \multicolumn{1}{c|}{\textbf{1.00}} &
  0.98 &
  \multicolumn{1}{c|}{1.00} &
  1.00 &
  \multicolumn{1}{c|}{1.00} &
  1.00 &
  1.00 \\
\multicolumn{1}{c|}{} &
  F1 Score &
  0.85 &
  \multicolumn{1}{c|}{0.89} &
  0.95 &
  \multicolumn{1}{c|}{\textbf{1.00}} &
  0.98 &
  \multicolumn{1}{c|}{1.00} &
  1.00 &
  \multicolumn{1}{c|}{1.00} &
  1.00 &
  1.00 \\
\multicolumn{1}{c|}{} &
  MCC &
  0.85 &
  \multicolumn{1}{c|}{0.88} &
  0.95 &
  \multicolumn{1}{c|}{\textbf{1.00}} &
  0.98 &
  \multicolumn{1}{c|}{1.00} &
  1.00 &
  \multicolumn{1}{c|}{1.00} &
  1.00 &
  1.00 \\ \hline
\multicolumn{1}{c|}{\multirow{3}{*}{Yale}} &
  Balanced Accuracy &
  0.14 &
  \multicolumn{1}{c|}{0.33} &
  0.25 &
  \multicolumn{1}{c|}{0.50} &
  0.39 &
  \multicolumn{1}{c|}{0.67} &
  0.37 &
  \multicolumn{1}{c|}{0.69} &
  0.53 &
  \textbf{0.70} \\
\multicolumn{1}{c|}{} &
  F1 Score &
  0.12 &
  \multicolumn{1}{c|}{0.31} &
  0.25 &
  \multicolumn{1}{c|}{0.50} &
  0.38 &
  \multicolumn{1}{c|}{0.67} &
  0.36 &
  \multicolumn{1}{c|}{\textbf{0.70}} &
  0.53 &
  0.66 \\
\multicolumn{1}{c|}{} &
  MCC &
  0.08 &
  \multicolumn{1}{c|}{0.27} &
  0.21 &
  \multicolumn{1}{c|}{0.47} &
  0.36 &
  \multicolumn{1}{c|}{0.66} &
  0.34 &
  \multicolumn{1}{c|}{\textbf{0.68}} &
  0.51 &
  0.68 \\ \hline
\multicolumn{1}{c|}{\multirow{3}{*}{USPS}} &
  Balanced Accuracy &
  0.72 &
  \multicolumn{1}{c|}{0.65} &
  0.90 &
  \multicolumn{1}{c|}{0.90} &
  0.91 &
  \multicolumn{1}{c|}{0.92} &
  0.92 &
  \multicolumn{1}{c|}{\textbf{0.93}} &
  0.92 &
  0.93 \\
\multicolumn{1}{c|}{} &
  F1 Score &
  0.71 &
  \multicolumn{1}{c|}{0.64} &
  0.90 &
  \multicolumn{1}{c|}{0.91} &
  0.91 &
  \multicolumn{1}{c|}{0.92} &
  0.92 &
  \multicolumn{1}{c|}{\textbf{0.93}} &
  0.92 &
  0.93 \\
\multicolumn{1}{c|}{} &
  MCC &
  0.72 &
  \multicolumn{1}{c|}{0.65} &
  0.90 &
  \multicolumn{1}{c|}{0.90} &
  0.91 &
  \multicolumn{1}{c|}{0.92} &
  0.92 &
  \multicolumn{1}{c|}{\textbf{0.93}} &
  0.92 &
  0.93 \\ \hline
\multicolumn{1}{c|}{\multirow{3}{*}{colon}} &
  Balanced Accuracy &
  0.70 &
  \multicolumn{1}{c|}{0.69} &
  0.69 &
  \multicolumn{1}{c|}{0.66} &
  \textbf{0.92} &
  \multicolumn{1}{c|}{0.82} &
  0.85 &
  \multicolumn{1}{c|}{0.74} &
  0.85 &
  0.88 \\
\multicolumn{1}{c|}{} &
  F1 Score &
  0.71 &
  \multicolumn{1}{c|}{0.68} &
  0.68 &
  \multicolumn{1}{c|}{0.66} &
  \textbf{0.89} &
  \multicolumn{1}{c|}{0.82} &
  0.83 &
  \multicolumn{1}{c|}{0.76} &
  0.83 &
  0.84 \\
\multicolumn{1}{c|}{} &
  MCC &
  0.42 &
  \multicolumn{1}{c|}{0.37} &
  0.37 &
  \multicolumn{1}{c|}{0.32} &
  \textbf{0.81} &
  \multicolumn{1}{c|}{0.65} &
  0.67 &
  \multicolumn{1}{c|}{0.53} &
  0.67 &
  0.72 \\ \hline
\multicolumn{1}{c|}{\multirow{3}{*}{GLIOMA}} &
  Balanced Accuracy &
  \textbf{0.61} &
  \multicolumn{1}{c|}{0.25} &
  0.30 &
  \multicolumn{1}{c|}{0.30} &
  0.30 &
  \multicolumn{1}{c|}{0.38} &
  0.60 &
  \multicolumn{1}{c|}{0.41} &
  0.59 &
  0.25 \\
\multicolumn{1}{c|}{} &
  F1 Score &
  0.57 &
  \multicolumn{1}{c|}{0.12} &
  0.24 &
  \multicolumn{1}{c|}{0.28} &
  0.26 &
  \multicolumn{1}{c|}{0.28} &
  \textbf{0.59} &
  \multicolumn{1}{c|}{0.35} &
  0.58 &
  0.13 \\
\multicolumn{1}{c|}{} &
  MCC &
  \textbf{0.50} &
  \multicolumn{1}{c|}{0.00} &
  0.03 &
  \multicolumn{1}{c|}{0.06} &
  0.10 &
  \multicolumn{1}{c|}{0.22} &
  0.47 &
  \multicolumn{1}{c|}{0.26} &
  0.46 &
  -0.03 \\ \hline
\multicolumn{1}{c|}{\multirow{3}{*}{lung}} &
  Balanced Accuracy &
  0.39 &
  \multicolumn{1}{c|}{0.47} &
  0.67 &
  \multicolumn{1}{c|}{0.89} &
  0.81 &
  \multicolumn{1}{c|}{\textbf{0.95}} &
  0.71 &
  \multicolumn{1}{c|}{0.87} &
  0.90 &
  0.81 \\
\multicolumn{1}{c|}{} &
  F1 Score &
  0.42 &
  \multicolumn{1}{c|}{0.50} &
  0.67 &
  \multicolumn{1}{c|}{0.88} &
  0.79 &
  \multicolumn{1}{c|}{\textbf{0.91}} &
  0.68 &
  \multicolumn{1}{c|}{0.86} &
  0.87 &
  0.83 \\
\multicolumn{1}{c|}{} &
  MCC &
  0.31 &
  \multicolumn{1}{c|}{0.61} &
  0.70 &
  \multicolumn{1}{c|}{0.80} &
  0.77 &
  \multicolumn{1}{c|}{\textbf{0.85}} &
  0.77 &
  \multicolumn{1}{c|}{0.77} &
  0.79 &
  0.80 \\ \hline
\multicolumn{1}{c|}{\multirow{3}{*}{lung\_small}} &
  Balanced Accuracy &
  0.49 &
  \multicolumn{1}{c|}{0.57} &
  0.76 &
  \multicolumn{1}{c|}{0.79} &
  0.82 &
  \multicolumn{1}{c|}{0.68} &
  0.79 &
  \multicolumn{1}{c|}{0.75} &
  0.82 &
  \textbf{0.93} \\
\multicolumn{1}{c|}{} &
  F1 Score &
  0.52 &
  \multicolumn{1}{c|}{0.55} &
  0.73 &
  \multicolumn{1}{c|}{0.71} &
  0.78 &
  \multicolumn{1}{c|}{0.65} &
  0.74 &
  \multicolumn{1}{c|}{0.69} &
  0.78 &
  \textbf{0.93} \\
\multicolumn{1}{c|}{} &
  MCC &
  0.51 &
  \multicolumn{1}{c|}{0.67} &
  0.72 &
  \multicolumn{1}{c|}{0.78} &
  0.84 &
  \multicolumn{1}{c|}{0.73} &
  0.79 &
  \multicolumn{1}{c|}{0.74} &
  0.84 &
  \textbf{0.90} \\ \hline
\multicolumn{1}{c|}{\multirow{3}{*}{lymphoma}} &
  Balanced Accuracy &
  0.22 &
  \multicolumn{1}{c|}{0.50} &
  0.58 &
  \multicolumn{1}{c|}{0.96} &
  0.78 &
  \multicolumn{1}{c|}{0.87} &
  0.90 &
  \multicolumn{1}{c|}{0.82} &
  0.81 &
  \textbf{0.98} \\
\multicolumn{1}{c|}{} &
  F1 Score &
  0.22 &
  \multicolumn{1}{c|}{0.47} &
  0.52 &
  \multicolumn{1}{c|}{0.91} &
  0.77 &
  \multicolumn{1}{c|}{0.83} &
  0.92 &
  \multicolumn{1}{c|}{0.78} &
  0.77 &
  \textbf{0.96} \\
\multicolumn{1}{c|}{} &
  MCC &
  0.33 &
  \multicolumn{1}{c|}{0.66} &
  0.44 &
  \multicolumn{1}{c|}{0.84} &
  0.81 &
  \multicolumn{1}{c|}{\textbf{0.91}} &
  0.91 &
  \multicolumn{1}{c|}{0.82} &
  0.78 &
  0.91 \\ \hline
\multicolumn{1}{c|}{\multirow{3}{*}{GISETTE}} &
  Balanced Accuracy &
  0.50 &
  \multicolumn{1}{c|}{0.49} &
  0.48 &
  \multicolumn{1}{c|}{0.47} &
  0.51 &
  \multicolumn{1}{c|}{\textbf{0.52}} &
  0.47 &
  \multicolumn{1}{c|}{0.50} &
  0.49 &
  0.50 \\
\multicolumn{1}{c|}{} &
  F1 Score &
  0.38 &
  \multicolumn{1}{c|}{0.49} &
  0.46 &
  \multicolumn{1}{c|}{0.44} &
  0.46 &
  \multicolumn{1}{c|}{\textbf{0.52}} &
  0.43 &
  \multicolumn{1}{c|}{0.45} &
  0.45 &
  0.46 \\
\multicolumn{1}{c|}{} &
  MCC &
  -0.01 &
  \multicolumn{1}{c|}{-0.01} &
  -0.05 &
  \multicolumn{1}{c|}{-0.07} &
  0.02 &
  \multicolumn{1}{c|}{\textbf{0.05}} &
  -0.08 &
  \multicolumn{1}{c|}{0.00} &
  -0.02 &
  0.00 \\ \hline
\multicolumn{1}{c|}{\multirow{3}{*}{Isolet}} &
  Balanced Accuracy &
  0.32 &
  \multicolumn{1}{c|}{0.51} &
  0.74 &
  \multicolumn{1}{c|}{0.78} &
  0.81 &
  \multicolumn{1}{c|}{0.82} &
  0.88 &
  \multicolumn{1}{c|}{0.83} &
  0.89 &
  \textbf{0.89} \\
\multicolumn{1}{c|}{} &
  F1 Score &
  0.26 &
  \multicolumn{1}{c|}{0.48} &
  0.73 &
  \multicolumn{1}{c|}{0.78} &
  0.81 &
  \multicolumn{1}{c|}{0.82} &
  0.88 &
  \multicolumn{1}{c|}{0.83} &
  0.89 &
  \textbf{0.89} \\
\multicolumn{1}{c|}{} &
  MCC &
  0.30 &
  \multicolumn{1}{c|}{0.49} &
  0.73 &
  \multicolumn{1}{c|}{0.77} &
  0.80 &
  \multicolumn{1}{c|}{0.81} &
  0.88 &
  \multicolumn{1}{c|}{0.82} &
  0.88 &
  \textbf{0.89} \\ \hline
\multicolumn{1}{c|}{\multirow{3}{*}{MADELON}} &
  Balanced Accuracy &
  0.59 &
  \multicolumn{1}{c|}{\textbf{0.59}} &
  0.58 &
  \multicolumn{1}{c|}{0.56} &
  0.55 &
  \multicolumn{1}{c|}{0.57} &
  0.54 &
  \multicolumn{1}{c|}{0.57} &
  0.57 &
  0.57 \\
\multicolumn{1}{c|}{} &
  F1 Score &
  0.59 &
  \multicolumn{1}{c|}{\textbf{0.59}} &
  0.58 &
  \multicolumn{1}{c|}{0.55} &
  0.55 &
  \multicolumn{1}{c|}{0.57} &
  0.54 &
  \multicolumn{1}{c|}{0.57} &
  0.57 &
  0.57 \\
\multicolumn{1}{c|}{} &
  MCC &
  0.18 &
  \multicolumn{1}{c|}{\textbf{0.18}} &
  0.16 &
  \multicolumn{1}{c|}{0.11} &
  0.10 &
  \multicolumn{1}{c|}{0.14} &
  0.09 &
  \multicolumn{1}{c|}{0.15} &
  0.14 &
  0.14 \\ \hline
\end{tabular}
}
\end{table}

\begin{table}[H]
\centering
\caption{Subset size-dependent, out-of-sample results using a KNN classifier. We use three evaluation metrics: the Balanced Accuracy score, the F1 score and the Matthews Correlation Coefficient. For each dataset, we boldly highlight the combination between feature selection schema and classifier producing the best out-of-sample result.}
\label{tab:local_bests_all_metrics_KNN}
\vspace{0.1in}
\scalebox{0.75}{
\begin{tabular}{cc|cccccccccc}
\hline
 &
   &
  \multicolumn{10}{c}{\textbf{KNN}} \\ \cline{3-12} 
 &
   &
  \multicolumn{2}{c|}{\textbf{10}} &
  \multicolumn{2}{c|}{\textbf{50}} &
  \multicolumn{2}{c|}{\textbf{100}} &
  \multicolumn{2}{c|}{\textbf{150}} &
  \multicolumn{2}{c}{\textbf{200}} \\ \cline{3-12} 
 &
   &
  \textbf{$\textrm{Inf-FS}_U$} &
  \multicolumn{1}{c|}{\textbf{TFS}} &
  \textbf{$\textrm{Inf-FS}_U$} &
  \multicolumn{1}{c|}{\textbf{TFS}} &
  \textbf{$\textrm{Inf-FS}_U$} &
  \multicolumn{1}{c|}{\textbf{TFS}} &
  \textbf{$\textrm{Inf-FS}_U$} &
  \multicolumn{1}{c|}{\textbf{TFS}} &
  \textbf{$\textrm{Inf-FS}_U$} &
  \textbf{TFS} \\ \hline
\multicolumn{1}{c|}{\multirow{3}{*}{PCMAC}} &
  Balanced Accuracy &
  0.52 &
  \multicolumn{1}{c|}{0.53} &
  0.57 &
  \multicolumn{1}{c|}{0.61} &
  0.61 &
  \multicolumn{1}{c|}{\textbf{0.62}} &
  0.61 &
  \multicolumn{1}{c|}{0.62} &
  0.61 &
  0.62 \\
\multicolumn{1}{c|}{} &
  F1 Score &
  0.52 &
  \multicolumn{1}{c|}{0.41} &
  0.57 &
  \multicolumn{1}{c|}{0.61} &
  0.61 &
  \multicolumn{1}{c|}{0.61} &
  0.59 &
  \multicolumn{1}{c|}{\textbf{0.63}} &
  0.60 &
  0.62 \\
\multicolumn{1}{c|}{} &
  MCC &
  0.05 &
  \multicolumn{1}{c|}{0.15} &
  0.15 &
  \multicolumn{1}{c|}{0.22} &
  0.23 &
  \multicolumn{1}{c|}{0.25} &
  0.18 &
  \multicolumn{1}{c|}{\textbf{0.26}} &
  0.21 &
  0.24 \\ \hline
\multicolumn{1}{c|}{\multirow{3}{*}{RELATHE}} &
  Balanced Accuracy &
  0.46 &
  \multicolumn{1}{c|}{0.46} &
  0.50 &
  \multicolumn{1}{c|}{\textbf{0.57}} &
  0.48 &
  \multicolumn{1}{c|}{0.49} &
  0.48 &
  \multicolumn{1}{c|}{0.49} &
  0.48 &
  0.49 \\
\multicolumn{1}{c|}{} &
  F1 Score &
  0.37 &
  \multicolumn{1}{c|}{0.37} &
  0.47 &
  \multicolumn{1}{c|}{\textbf{0.54}} &
  0.46 &
  \multicolumn{1}{c|}{0.46} &
  0.45 &
  \multicolumn{1}{c|}{0.45} &
  0.48 &
  0.46 \\
\multicolumn{1}{c|}{} &
  MCC &
  -0.10 &
  \multicolumn{1}{c|}{-0.10} &
  0.00 &
  \multicolumn{1}{c|}{\textbf{0.17}} &
  -0.04 &
  \multicolumn{1}{c|}{-0.03} &
  -0.09 &
  \multicolumn{1}{c|}{-0.10} &
  -0.03 &
  -0.05 \\ \hline
\multicolumn{1}{c|}{\multirow{3}{*}{COIL20}} &
  Balanced Accuracy &
  0.70 &
  \multicolumn{1}{c|}{0.82} &
  0.86 &
  \multicolumn{1}{c|}{\textbf{0.93}} &
  0.93 &
  \multicolumn{1}{c|}{0.93} &
  0.93 &
  \multicolumn{1}{c|}{0.93} &
  0.93 &
  0.93 \\
\multicolumn{1}{c|}{} &
  F1 Score &
  0.67 &
  \multicolumn{1}{c|}{0.81} &
  0.85 &
  \multicolumn{1}{c|}{0.93} &
  0.92 &
  \multicolumn{1}{c|}{0.93} &
  0.96 &
  \multicolumn{1}{c|}{0.94} &
  \textbf{0.97} &
  0.93 \\
\multicolumn{1}{c|}{} &
  MCC &
  0.69 &
  \multicolumn{1}{c|}{0.81} &
  0.85 &
  \multicolumn{1}{c|}{0.93} &
  0.92 &
  \multicolumn{1}{c|}{0.93} &
  0.95 &
  \multicolumn{1}{c|}{0.94} &
  \textbf{0.97} &
  0.93 \\ \hline
\multicolumn{1}{c|}{\multirow{3}{*}{ORL}} &
  Balanced Accuracy &
  0.38 &
  \multicolumn{1}{c|}{0.52} &
  0.52 &
  \multicolumn{1}{c|}{\textbf{0.77}} &
  0.62 &
  \multicolumn{1}{c|}{0.70} &
  0.62 &
  \multicolumn{1}{c|}{0.70} &
  0.62 &
  0.70 \\
\multicolumn{1}{c|}{} &
  F1 Score &
  0.38 &
  \multicolumn{1}{c|}{0.49} &
  0.49 &
  \multicolumn{1}{c|}{0.75} &
  0.61 &
  \multicolumn{1}{c|}{0.68} &
  0.73 &
  \multicolumn{1}{c|}{0.69} &
  0.69 &
  \textbf{0.76} \\
\multicolumn{1}{c|}{} &
  MCC &
  0.37 &
  \multicolumn{1}{c|}{0.52} &
  0.52 &
  \multicolumn{1}{c|}{0.76} &
  0.62 &
  \multicolumn{1}{c|}{0.69} &
  0.73 &
  \multicolumn{1}{c|}{0.70} &
  0.72 &
  \textbf{0.77} \\ \hline
\multicolumn{1}{c|}{\multirow{3}{*}{warpAR10P}} &
  Balanced Accuracy &
  0.36 &
  \multicolumn{1}{c|}{0.30} &
  0.36 &
  \multicolumn{1}{c|}{\textbf{0.51}} &
  0.43 &
  \multicolumn{1}{c|}{0.46} &
  0.43 &
  \multicolumn{1}{c|}{0.46} &
  0.43 &
  0.46 \\
\multicolumn{1}{c|}{} &
  F1 Score &
  0.32 &
  \multicolumn{1}{c|}{0.26} &
  0.32 &
  \multicolumn{1}{c|}{\textbf{0.50}} &
  0.41 &
  \multicolumn{1}{c|}{0.48} &
  0.27 &
  \multicolumn{1}{c|}{0.37} &
  0.37 &
  0.47 \\
\multicolumn{1}{c|}{} &
  MCC &
  0.29 &
  \multicolumn{1}{c|}{0.24} &
  0.30 &
  \multicolumn{1}{c|}{\textbf{0.47}} &
  0.38 &
  \multicolumn{1}{c|}{0.41} &
  0.25 &
  \multicolumn{1}{c|}{0.32} &
  0.37 &
  0.44 \\ \hline
\multicolumn{1}{c|}{\multirow{3}{*}{warpPIE10P}} &
  Balanced Accuracy &
  0.83 &
  \multicolumn{1}{c|}{0.72} &
  0.86 &
  \multicolumn{1}{c|}{0.91} &
  0.92 &
  \multicolumn{1}{c|}{\textbf{0.97}} &
  0.92 &
  \multicolumn{1}{c|}{0.97} &
  0.92 &
  0.97 \\
\multicolumn{1}{c|}{} &
  F1 Score &
  0.83 &
  \multicolumn{1}{c|}{0.69} &
  0.86 &
  \multicolumn{1}{c|}{0.91} &
  0.92 &
  \multicolumn{1}{c|}{\textbf{0.97}} &
  0.89 &
  \multicolumn{1}{c|}{0.92} &
  0.89 &
  0.95 \\
\multicolumn{1}{c|}{} &
  MCC &
  0.81 &
  \multicolumn{1}{c|}{0.69} &
  0.84 &
  \multicolumn{1}{c|}{0.90} &
  0.91 &
  \multicolumn{1}{c|}{\textbf{0.97}} &
  0.88 &
  \multicolumn{1}{c|}{0.91} &
  0.88 &
  0.95 \\ \hline
\multicolumn{1}{c|}{\multirow{3}{*}{Yale}} &
  Balanced Accuracy &
  0.14 &
  \multicolumn{1}{c|}{\textbf{0.42}} &
  0.28 &
  \multicolumn{1}{c|}{0.41} &
  0.26 &
  \multicolumn{1}{c|}{0.42} &
  0.26 &
  \multicolumn{1}{c|}{0.42} &
  0.26 &
  0.42 \\
\multicolumn{1}{c|}{} &
  F1 Score &
  0.12 &
  \multicolumn{1}{c|}{0.40} &
  0.27 &
  \multicolumn{1}{c|}{0.40} &
  0.23 &
  \multicolumn{1}{c|}{0.43} &
  0.44 &
  \multicolumn{1}{c|}{0.38} &
  0.32 &
  \textbf{0.49} \\
\multicolumn{1}{c|}{} &
  MCC &
  0.08 &
  \multicolumn{1}{c|}{0.39} &
  0.23 &
  \multicolumn{1}{c|}{0.36} &
  0.23 &
  \multicolumn{1}{c|}{0.39} &
  0.41 &
  \multicolumn{1}{c|}{0.34} &
  0.33 &
  \textbf{0.48} \\ \hline
\multicolumn{1}{c|}{\multirow{3}{*}{USPS}} &
  Balanced Accuracy &
  0.78 &
  \multicolumn{1}{c|}{0.77} &
  0.94 &
  \multicolumn{1}{c|}{0.94} &
  \textbf{0.96} &
  \multicolumn{1}{c|}{0.95} &
  0.96 &
  \multicolumn{1}{c|}{0.95} &
  0.96 &
  0.95 \\
\multicolumn{1}{c|}{} &
  F1 Score &
  0.78 &
  \multicolumn{1}{c|}{0.77} &
  0.94 &
  \multicolumn{1}{c|}{0.94} &
  \textbf{0.96} &
  \multicolumn{1}{c|}{0.95} &
  0.96 &
  \multicolumn{1}{c|}{0.95} &
  0.95 &
  0.95 \\
\multicolumn{1}{c|}{} &
  MCC &
  0.78 &
  \multicolumn{1}{c|}{0.78} &
  0.94 &
  \multicolumn{1}{c|}{0.94} &
  \textbf{0.96} &
  \multicolumn{1}{c|}{0.95} &
  0.96 &
  \multicolumn{1}{c|}{0.95} &
  0.95 &
  0.95 \\ \hline
\multicolumn{1}{c|}{\multirow{3}{*}{colon}} &
  Balanced Accuracy &
  0.77 &
  \multicolumn{1}{c|}{0.82} &
  \textbf{0.89} &
  \multicolumn{1}{c|}{0.77} &
  0.89 &
  \multicolumn{1}{c|}{0.85} &
  0.89 &
  \multicolumn{1}{c|}{0.85} &
  0.89 &
  0.85 \\
\multicolumn{1}{c|}{} &
  F1 Score &
  0.77 &
  \multicolumn{1}{c|}{0.82} &
  0.89 &
  \multicolumn{1}{c|}{0.77} &
  0.89 &
  \multicolumn{1}{c|}{0.83} &
  \textbf{1.00} &
  \multicolumn{1}{c|}{0.71} &
  0.83 &
  0.77 \\
\multicolumn{1}{c|}{} &
  MCC &
  0.55 &
  \multicolumn{1}{c|}{0.65} &
  0.77 &
  \multicolumn{1}{c|}{0.55} &
  0.77 &
  \multicolumn{1}{c|}{0.67} &
  \textbf{1.00} &
  \multicolumn{1}{c|}{0.42} &
  0.67 &
  0.55 \\ \hline
\multicolumn{1}{c|}{\multirow{3}{*}{GLIOMA}} &
  Balanced Accuracy &
  0.24 &
  \multicolumn{1}{c|}{0.10} &
  0.40 &
  \multicolumn{1}{c|}{0.40} &
  0.42 &
  \multicolumn{1}{c|}{\textbf{0.62}} &
  0.42 &
  \multicolumn{1}{c|}{0.62} &
  0.42 &
  0.62 \\
\multicolumn{1}{c|}{} &
  F1 Score &
  0.19 &
  \multicolumn{1}{c|}{0.06} &
  0.39 &
  \multicolumn{1}{c|}{0.40} &
  0.40 &
  \multicolumn{1}{c|}{0.49} &
  \textbf{0.52} &
  \multicolumn{1}{c|}{0.49} &
  0.50 &
  0.49 \\
\multicolumn{1}{c|}{} &
  MCC &
  -0.04 &
  \multicolumn{1}{c|}{-0.31} &
  0.26 &
  \multicolumn{1}{c|}{0.27} &
  0.30 &
  \multicolumn{1}{c|}{\textbf{0.48}} &
  0.46 &
  \multicolumn{1}{c|}{0.48} &
  0.46 &
  0.48 \\ \hline
\multicolumn{1}{c|}{\multirow{3}{*}{lung}} &
  Balanced Accuracy &
  0.33 &
  \multicolumn{1}{c|}{0.51} &
  0.65 &
  \multicolumn{1}{c|}{\textbf{0.79}} &
  0.71 &
  \multicolumn{1}{c|}{0.65} &
  0.71 &
  \multicolumn{1}{c|}{0.65} &
  0.71 &
  0.65 \\
\multicolumn{1}{c|}{} &
  F1 Score &
  0.35 &
  \multicolumn{1}{c|}{0.52} &
  0.65 &
  \multicolumn{1}{c|}{\textbf{0.84}} &
  0.70 &
  \multicolumn{1}{c|}{0.67} &
  0.71 &
  \multicolumn{1}{c|}{0.69} &
  0.81 &
  0.84 \\
\multicolumn{1}{c|}{} &
  MCC &
  0.24 &
  \multicolumn{1}{c|}{0.56} &
  0.70 &
  \multicolumn{1}{c|}{\textbf{0.83}} &
  0.77 &
  \multicolumn{1}{c|}{0.72} &
  0.80 &
  \multicolumn{1}{c|}{0.76} &
  0.77 &
  0.83 \\ \hline
\multicolumn{1}{c|}{\multirow{3}{*}{lung\_small}} &
  Balanced Accuracy &
  0.57 &
  \multicolumn{1}{c|}{0.61} &
  0.80 &
  \multicolumn{1}{c|}{0.87} &
  0.82 &
  \multicolumn{1}{c|}{\textbf{0.90}} &
  0.82 &
  \multicolumn{1}{c|}{0.90} &
  0.82 &
  0.90 \\
\multicolumn{1}{c|}{} &
  F1 Score &
  0.55 &
  \multicolumn{1}{c|}{0.57} &
  0.77 &
  \multicolumn{1}{c|}{0.87} &
  0.80 &
  \multicolumn{1}{c|}{0.85} &
  \textbf{0.89} &
  \multicolumn{1}{c|}{0.85} &
  0.85 &
  0.72 \\
\multicolumn{1}{c|}{} &
  MCC &
  0.51 &
  \multicolumn{1}{c|}{0.64} &
  0.78 &
  \multicolumn{1}{c|}{0.84} &
  0.84 &
  \multicolumn{1}{c|}{0.84} &
  \textbf{0.90} &
  \multicolumn{1}{c|}{0.84} &
  0.84 &
  0.73 \\ \hline
\multicolumn{1}{c|}{\multirow{3}{*}{lymphoma}} &
  Balanced Accuracy &
  0.44 &
  \multicolumn{1}{c|}{0.50} &
  0.60 &
  \multicolumn{1}{c|}{\textbf{0.74}} &
  0.69 &
  \multicolumn{1}{c|}{0.69} &
  0.69 &
  \multicolumn{1}{c|}{0.69} &
  0.69 &
  0.69 \\
\multicolumn{1}{c|}{} &
  F1 Score &
  0.45 &
  \multicolumn{1}{c|}{0.49} &
  0.61 &
  \multicolumn{1}{c|}{0.70} &
  0.67 &
  \multicolumn{1}{c|}{0.65} &
  \textbf{0.75} &
  \multicolumn{1}{c|}{0.72} &
  0.65 &
  0.70 \\
\multicolumn{1}{c|}{} &
  MCC &
  0.49 &
  \multicolumn{1}{c|}{0.66} &
  0.76 &
  \multicolumn{1}{c|}{0.86} &
  0.86 &
  \multicolumn{1}{c|}{0.86} &
  0.81 &
  \multicolumn{1}{c|}{\textbf{0.91}} &
  0.86 &
  0.86 \\ \hline
\multicolumn{1}{c|}{\multirow{3}{*}{GISETTE}} &
  Balanced Accuracy &
  0.49 &
  \multicolumn{1}{c|}{0.51} &
  0.52 &
  \multicolumn{1}{c|}{\textbf{0.54}} &
  0.50 &
  \multicolumn{1}{c|}{0.51} &
  0.50 &
  \multicolumn{1}{c|}{0.51} &
  0.50 &
  0.51 \\
\multicolumn{1}{c|}{} &
  F1 Score &
  0.38 &
  \multicolumn{1}{c|}{0.50} &
  0.49 &
  \multicolumn{1}{c|}{\textbf{0.52}} &
  0.40 &
  \multicolumn{1}{c|}{0.43} &
  0.34 &
  \multicolumn{1}{c|}{0.48} &
  0.34 &
  0.47 \\
\multicolumn{1}{c|}{} &
  MCC &
  -0.03 &
  \multicolumn{1}{c|}{0.02} &
  0.04 &
  \multicolumn{1}{c|}{\textbf{0.09}} &
  -0.01 &
  \multicolumn{1}{c|}{0.03} &
  -0.03 &
  \multicolumn{1}{c|}{0.08} &
  -0.03 &
  -0.01 \\ \hline
\multicolumn{1}{c|}{\multirow{3}{*}{Isolet}} &
  Balanced Accuracy &
  0.32 &
  \multicolumn{1}{c|}{0.49} &
  0.72 &
  \multicolumn{1}{c|}{0.73} &
  0.78 &
  \multicolumn{1}{c|}{\textbf{0.78}} &
  0.78 &
  \multicolumn{1}{c|}{0.78} &
  0.78 &
  0.78 \\
\multicolumn{1}{c|}{} &
  F1 Score &
  0.31 &
  \multicolumn{1}{c|}{0.48} &
  0.72 &
  \multicolumn{1}{c|}{0.73} &
  0.78 &
  \multicolumn{1}{c|}{0.77} &
  \textbf{0.83} &
  \multicolumn{1}{c|}{0.81} &
  0.82 &
  0.83 \\
\multicolumn{1}{c|}{} &
  MCC &
  0.29 &
  \multicolumn{1}{c|}{0.47} &
  0.71 &
  \multicolumn{1}{c|}{0.72} &
  0.78 &
  \multicolumn{1}{c|}{0.77} &
  0.82 &
  \multicolumn{1}{c|}{0.81} &
  0.81 &
  \textbf{0.83} \\ \hline
\multicolumn{1}{c|}{\multirow{3}{*}{MADELON}} &
  Balanced Accuracy &
  0.61 &
  \multicolumn{1}{c|}{\textbf{0.78}} &
  0.58 &
  \multicolumn{1}{c|}{0.74} &
  0.64 &
  \multicolumn{1}{c|}{0.66} &
  0.64 &
  \multicolumn{1}{c|}{0.66} &
  0.64 &
  0.66 \\
\multicolumn{1}{c|}{} &
  F1 Score &
  0.61 &
  \multicolumn{1}{c|}{\textbf{0.78}} &
  0.58 &
  \multicolumn{1}{c|}{0.74} &
  0.64 &
  \multicolumn{1}{c|}{0.66} &
  0.62 &
  \multicolumn{1}{c|}{0.64} &
  0.57 &
  0.62 \\
\multicolumn{1}{c|}{} &
  MCC &
  0.23 &
  \multicolumn{1}{c|}{\textbf{0.56}} &
  0.16 &
  \multicolumn{1}{c|}{0.48} &
  0.29 &
  \multicolumn{1}{c|}{0.32} &
  0.24 &
  \multicolumn{1}{c|}{0.29} &
  0.14 &
  0.26 \\ \hline
\end{tabular}
}
\end{table}

\begin{table}[H]
\centering
\caption{Subset size-dependent, out-of-sample results using a Decision Tree classifier. We use three evaluation metrics: the Balanced Accuracy score, the F1 score and the Matthews Correlation Coefficient. For each dataset, we boldly highlight the combination between feature selection schema and classifier producing the best out-of-sample result.}
\label{tab:local_bests_all_metrics_DecisionTree}
\vspace{0.1in}
\scalebox{0.75}{
\begin{tabular}{cc|cccccccccc}
\hline
 &
   &
  \multicolumn{10}{c}{\textbf{Decision Tree}} \\ \cline{3-12} 
 &
   &
  \multicolumn{2}{c|}{\textbf{10}} &
  \multicolumn{2}{c|}{\textbf{50}} &
  \multicolumn{2}{c|}{\textbf{100}} &
  \multicolumn{2}{c|}{\textbf{150}} &
  \multicolumn{2}{c}{\textbf{200}} \\ \cline{3-12} 
 &
   &
  \textbf{$\textrm{Inf-FS}_U$} &
  \multicolumn{1}{c|}{\textbf{TFS}} &
  \textbf{$\textrm{Inf-FS}_U$} &
  \multicolumn{1}{c|}{\textbf{TFS}} &
  \textbf{$\textrm{Inf-FS}_U$} &
  \multicolumn{1}{c|}{\textbf{TFS}} &
  \textbf{$\textrm{Inf-FS}_U$} &
  \multicolumn{1}{c|}{\textbf{TFS}} &
  \textbf{$\textrm{Inf-FS}_U$} &
  \textbf{TFS} \\ \hline
\multicolumn{1}{c|}{\multirow{3}{*}{PCMAC}} &
  Balanced Accuracy &
  0.53 &
  \multicolumn{1}{c|}{0.50} &
  0.56 &
  \multicolumn{1}{c|}{0.69} &
  0.58 &
  \multicolumn{1}{c|}{\textbf{0.71}} &
  0.58 &
  \multicolumn{1}{c|}{0.71} &
  0.58 &
  0.71 \\
\multicolumn{1}{c|}{} &
  F1 Score &
  0.53 &
  \multicolumn{1}{c|}{0.35} &
  0.56 &
  \multicolumn{1}{c|}{0.69} &
  0.58 &
  \multicolumn{1}{c|}{0.71} &
  0.57 &
  \multicolumn{1}{c|}{0.68} &
  0.60 &
  \textbf{0.73} \\
\multicolumn{1}{c|}{} &
  MCC &
  0.06 &
  \multicolumn{1}{c|}{-0.01} &
  0.13 &
  \multicolumn{1}{c|}{0.38} &
  0.17 &
  \multicolumn{1}{c|}{0.42} &
  0.14 &
  \multicolumn{1}{c|}{0.36} &
  0.21 &
  \textbf{0.46} \\ \hline
\multicolumn{1}{c|}{\multirow{3}{*}{RELATHE}} &
  Balanced Accuracy &
  0.49 &
  \multicolumn{1}{c|}{0.50} &
  0.51 &
  \multicolumn{1}{c|}{\textbf{0.51}} &
  0.49 &
  \multicolumn{1}{c|}{0.42} &
  0.49 &
  \multicolumn{1}{c|}{0.42} &
  0.49 &
  0.42 \\
\multicolumn{1}{c|}{} &
  F1 Score &
  0.35 &
  \multicolumn{1}{c|}{0.34} &
  0.41 &
  \multicolumn{1}{c|}{0.47} &
  0.46 &
  \multicolumn{1}{c|}{0.41} &
  0.45 &
  \multicolumn{1}{c|}{\textbf{0.48}} &
  0.48 &
  0.44 \\
\multicolumn{1}{c|}{} &
  MCC &
  -0.02 &
  \multicolumn{1}{c|}{-0.02} &
  \textbf{0.03} &
  \multicolumn{1}{c|}{0.02} &
  -0.01 &
  \multicolumn{1}{c|}{-0.18} &
  -0.04 &
  \multicolumn{1}{c|}{0.02} &
  -0.04 &
  0.03 \\ \hline
\multicolumn{1}{c|}{\multirow{3}{*}{COIL20}} &
  Balanced Accuracy &
  0.68 &
  \multicolumn{1}{c|}{0.81} &
  0.83 &
  \multicolumn{1}{c|}{0.89} &
  0.85 &
  \multicolumn{1}{c|}{\textbf{0.90}} &
  0.85 &
  \multicolumn{1}{c|}{0.90} &
  0.85 &
  0.90 \\
\multicolumn{1}{c|}{} &
  F1 Score &
  0.67 &
  \multicolumn{1}{c|}{0.81} &
  0.83 &
  \multicolumn{1}{c|}{0.89} &
  0.85 &
  \multicolumn{1}{c|}{\textbf{0.90}} &
  0.89 &
  \multicolumn{1}{c|}{0.90} &
  0.90 &
  0.90 \\
\multicolumn{1}{c|}{} &
  MCC &
  0.67 &
  \multicolumn{1}{c|}{0.80} &
  0.82 &
  \multicolumn{1}{c|}{0.89} &
  0.84 &
  \multicolumn{1}{c|}{\textbf{0.90}} &
  0.89 &
  \multicolumn{1}{c|}{0.90} &
  0.90 &
  0.90 \\ \hline
\multicolumn{1}{c|}{\multirow{3}{*}{ORL}} &
  Balanced Accuracy &
  0.36 &
  \multicolumn{1}{c|}{0.39} &
  0.42 &
  \multicolumn{1}{c|}{0.48} &
  0.49 &
  \multicolumn{1}{c|}{\textbf{0.54}} &
  0.49 &
  \multicolumn{1}{c|}{0.54} &
  0.49 &
  0.54 \\
\multicolumn{1}{c|}{} &
  F1 Score &
  0.33 &
  \multicolumn{1}{c|}{0.37} &
  0.40 &
  \multicolumn{1}{c|}{0.46} &
  0.49 &
  \multicolumn{1}{c|}{0.52} &
  0.57 &
  \multicolumn{1}{c|}{0.58} &
  0.49 &
  \textbf{0.60} \\
\multicolumn{1}{c|}{} &
  MCC &
  0.34 &
  \multicolumn{1}{c|}{0.38} &
  0.41 &
  \multicolumn{1}{c|}{0.47} &
  0.48 &
  \multicolumn{1}{c|}{0.53} &
  0.58 &
  \multicolumn{1}{c|}{0.60} &
  0.48 &
  \textbf{0.61} \\ \hline
\multicolumn{1}{c|}{\multirow{3}{*}{warpAR10P}} &
  Balanced Accuracy &
  0.37 &
  \multicolumn{1}{c|}{0.33} &
  0.46 &
  \multicolumn{1}{c|}{\textbf{0.59}} &
  0.55 &
  \multicolumn{1}{c|}{0.59} &
  0.55 &
  \multicolumn{1}{c|}{0.59} &
  0.55 &
  0.59 \\
\multicolumn{1}{c|}{} &
  F1 Score &
  0.35 &
  \multicolumn{1}{c|}{0.33} &
  0.44 &
  \multicolumn{1}{c|}{0.60} &
  0.52 &
  \multicolumn{1}{c|}{0.61} &
  0.41 &
  \multicolumn{1}{c|}{0.64} &
  0.67 &
  \textbf{0.81} \\
\multicolumn{1}{c|}{} &
  MCC &
  0.29 &
  \multicolumn{1}{c|}{0.27} &
  0.41 &
  \multicolumn{1}{c|}{0.55} &
  0.49 &
  \multicolumn{1}{c|}{0.55} &
  0.35 &
  \multicolumn{1}{c|}{0.61} &
  0.63 &
  \textbf{0.78} \\ \hline
\multicolumn{1}{c|}{\multirow{3}{*}{warpPIE10P}} &
  Balanced Accuracy &
  0.74 &
  \multicolumn{1}{c|}{0.74} &
  0.80 &
  \multicolumn{1}{c|}{0.73} &
  0.77 &
  \multicolumn{1}{c|}{\textbf{0.85}} &
  0.77 &
  \multicolumn{1}{c|}{0.85} &
  0.77 &
  0.85 \\
\multicolumn{1}{c|}{} &
  F1 Score &
  0.74 &
  \multicolumn{1}{c|}{0.75} &
  0.78 &
  \multicolumn{1}{c|}{0.73} &
  0.78 &
  \multicolumn{1}{c|}{0.85} &
  0.75 &
  \multicolumn{1}{c|}{\textbf{0.87}} &
  0.76 &
  0.79 \\
\multicolumn{1}{c|}{} &
  MCC &
  0.72 &
  \multicolumn{1}{c|}{0.72} &
  0.78 &
  \multicolumn{1}{c|}{0.71} &
  0.76 &
  \multicolumn{1}{c|}{0.84} &
  0.74 &
  \multicolumn{1}{c|}{\textbf{0.86}} &
  0.74 &
  0.79 \\ \hline
\multicolumn{1}{c|}{\multirow{3}{*}{Yale}} &
  Balanced Accuracy &
  0.17 &
  \multicolumn{1}{c|}{0.31} &
  0.26 &
  \multicolumn{1}{c|}{0.34} &
  0.39 &
  \multicolumn{1}{c|}{\textbf{0.42}} &
  0.39 &
  \multicolumn{1}{c|}{0.42} &
  0.39 &
  0.42 \\
\multicolumn{1}{c|}{} &
  F1 Score &
  0.15 &
  \multicolumn{1}{c|}{0.31} &
  0.25 &
  \multicolumn{1}{c|}{0.34} &
  0.39 &
  \multicolumn{1}{c|}{0.41} &
  0.47 &
  \multicolumn{1}{c|}{0.43} &
  0.45 &
  \textbf{0.53} \\
\multicolumn{1}{c|}{} &
  MCC &
  0.12 &
  \multicolumn{1}{c|}{0.26} &
  0.21 &
  \multicolumn{1}{c|}{0.30} &
  0.36 &
  \multicolumn{1}{c|}{0.38} &
  \textbf{0.49} &
  \multicolumn{1}{c|}{0.38} &
  0.40 &
  0.49 \\ \hline
\multicolumn{1}{c|}{\multirow{3}{*}{USPS}} &
  Balanced Accuracy &
  0.73 &
  \multicolumn{1}{c|}{0.72} &
  0.84 &
  \multicolumn{1}{c|}{0.85} &
  0.85 &
  \multicolumn{1}{c|}{\textbf{0.86}} &
  0.85 &
  \multicolumn{1}{c|}{0.86} &
  0.85 &
  0.86 \\
\multicolumn{1}{c|}{} &
  F1 Score &
  0.73 &
  \multicolumn{1}{c|}{0.72} &
  0.84 &
  \multicolumn{1}{c|}{0.85} &
  0.85 &
  \multicolumn{1}{c|}{0.86} &
  0.86 &
  \multicolumn{1}{c|}{0.86} &
  \textbf{0.88} &
  0.87 \\
\multicolumn{1}{c|}{} &
  MCC &
  0.73 &
  \multicolumn{1}{c|}{0.72} &
  0.84 &
  \multicolumn{1}{c|}{0.85} &
  0.85 &
  \multicolumn{1}{c|}{0.86} &
  0.86 &
  \multicolumn{1}{c|}{0.86} &
  \textbf{0.88} &
  0.87 \\ \hline
\multicolumn{1}{c|}{\multirow{3}{*}{colon}} &
  Balanced Accuracy &
  0.61 &
  \multicolumn{1}{c|}{0.64} &
  0.82 &
  \multicolumn{1}{c|}{0.74} &
  0.76 &
  \multicolumn{1}{c|}{\textbf{0.92}} &
  0.76 &
  \multicolumn{1}{c|}{0.92} &
  0.76 &
  0.92 \\
\multicolumn{1}{c|}{} &
  F1 Score &
  0.58 &
  \multicolumn{1}{c|}{0.64} &
  0.82 &
  \multicolumn{1}{c|}{0.76} &
  0.73 &
  \multicolumn{1}{c|}{\textbf{0.89}} &
  0.89 &
  \multicolumn{1}{c|}{0.79} &
  0.83 &
  0.79 \\
\multicolumn{1}{c|}{} &
  MCC &
  0.21 &
  \multicolumn{1}{c|}{0.45} &
  0.65 &
  \multicolumn{1}{c|}{0.53} &
  0.51 &
  \multicolumn{1}{c|}{\textbf{0.81}} &
  0.77 &
  \multicolumn{1}{c|}{0.65} &
  0.67 &
  0.65 \\ \hline
\multicolumn{1}{c|}{\multirow{3}{*}{GLIOMA}} &
  Balanced Accuracy &
  0.34 &
  \multicolumn{1}{c|}{\textbf{0.61}} &
  0.36 &
  \multicolumn{1}{c|}{0.31} &
  0.35 &
  \multicolumn{1}{c|}{0.44} &
  0.35 &
  \multicolumn{1}{c|}{0.44} &
  0.35 &
  0.44 \\
\multicolumn{1}{c|}{} &
  F1 Score &
  0.37 &
  \multicolumn{1}{c|}{\textbf{0.54}} &
  0.25 &
  \multicolumn{1}{c|}{0.18} &
  0.34 &
  \multicolumn{1}{c|}{0.39} &
  0.22 &
  \multicolumn{1}{c|}{0.22} &
  0.19 &
  0.33 \\
\multicolumn{1}{c|}{} &
  MCC &
  0.19 &
  \multicolumn{1}{c|}{\textbf{0.41}} &
  0.06 &
  \multicolumn{1}{c|}{0.10} &
  0.19 &
  \multicolumn{1}{c|}{0.21} &
  0.19 &
  \multicolumn{1}{c|}{0.14} &
  0.08 &
  0.33 \\ \hline
\multicolumn{1}{c|}{\multirow{3}{*}{lung}} &
  Balanced Accuracy &
  0.44 &
  \multicolumn{1}{c|}{0.70} &
  0.75 &
  \multicolumn{1}{c|}{0.71} &
  \textbf{0.87} &
  \multicolumn{1}{c|}{0.70} &
  0.87 &
  \multicolumn{1}{c|}{0.70} &
  0.87 &
  0.70 \\
\multicolumn{1}{c|}{} &
  F1 Score &
  0.44 &
  \multicolumn{1}{c|}{0.70} &
  0.67 &
  \multicolumn{1}{c|}{0.75} &
  0.83 &
  \multicolumn{1}{c|}{0.66} &
  \textbf{0.88} &
  \multicolumn{1}{c|}{0.73} &
  0.75 &
  0.75 \\
\multicolumn{1}{c|}{} &
  MCC &
  0.39 &
  \multicolumn{1}{c|}{0.64} &
  0.64 &
  \multicolumn{1}{c|}{0.69} &
  0.78 &
  \multicolumn{1}{c|}{0.69} &
  \textbf{0.79} &
  \multicolumn{1}{c|}{0.71} &
  0.67 &
  0.70 \\ \hline
\multicolumn{1}{c|}{\multirow{3}{*}{lung\_small}} &
  Balanced Accuracy &
  0.46 &
  \multicolumn{1}{c|}{0.42} &
  0.58 &
  \multicolumn{1}{c|}{\textbf{0.63}} &
  0.47 &
  \multicolumn{1}{c|}{0.57} &
  0.47 &
  \multicolumn{1}{c|}{0.57} &
  0.47 &
  0.57 \\
\multicolumn{1}{c|}{} &
  F1 Score &
  0.42 &
  \multicolumn{1}{c|}{0.37} &
  0.47 &
  \multicolumn{1}{c|}{0.58} &
  0.42 &
  \multicolumn{1}{c|}{0.48} &
  0.40 &
  \multicolumn{1}{c|}{\textbf{0.63}} &
  0.52 &
  0.37 \\
\multicolumn{1}{c|}{} &
  MCC &
  0.45 &
  \multicolumn{1}{c|}{0.50} &
  \textbf{0.58} &
  \multicolumn{1}{c|}{0.57} &
  0.40 &
  \multicolumn{1}{c|}{0.47} &
  0.47 &
  \multicolumn{1}{c|}{0.55} &
  0.44 &
  0.41 \\ \hline
\multicolumn{1}{c|}{\multirow{3}{*}{lymphoma}} &
  Balanced Accuracy &
  0.20 &
  \multicolumn{1}{c|}{\textbf{0.69}} &
  0.45 &
  \multicolumn{1}{c|}{0.55} &
  0.45 &
  \multicolumn{1}{c|}{0.44} &
  0.45 &
  \multicolumn{1}{c|}{0.44} &
  0.45 &
  0.44 \\
\multicolumn{1}{c|}{} &
  F1 Score &
  0.17 &
  \multicolumn{1}{c|}{\textbf{0.67}} &
  0.37 &
  \multicolumn{1}{c|}{0.45} &
  0.38 &
  \multicolumn{1}{c|}{0.39} &
  0.59 &
  \multicolumn{1}{c|}{0.49} &
  0.46 &
  0.49 \\
\multicolumn{1}{c|}{} &
  MCC &
  0.18 &
  \multicolumn{1}{c|}{\textbf{0.64}} &
  0.50 &
  \multicolumn{1}{c|}{0.57} &
  0.44 &
  \multicolumn{1}{c|}{0.50} &
  0.62 &
  \multicolumn{1}{c|}{0.59} &
  0.57 &
  0.60 \\ \hline
\multicolumn{1}{c|}{\multirow{3}{*}{GISETTE}} &
  Balanced Accuracy &
  \textbf{0.52} &
  \multicolumn{1}{c|}{0.50} &
  0.44 &
  \multicolumn{1}{c|}{0.52} &
  0.48 &
  \multicolumn{1}{c|}{0.47} &
  0.48 &
  \multicolumn{1}{c|}{0.47} &
  0.48 &
  0.47 \\
\multicolumn{1}{c|}{} &
  F1 Score &
  0.45 &
  \multicolumn{1}{c|}{\textbf{0.50}} &
  0.41 &
  \multicolumn{1}{c|}{0.49} &
  0.48 &
  \multicolumn{1}{c|}{0.46} &
  0.46 &
  \multicolumn{1}{c|}{0.48} &
  0.49 &
  0.44 \\
\multicolumn{1}{c|}{} &
  MCC &
  \textbf{0.05} &
  \multicolumn{1}{c|}{0.00} &
  -0.14 &
  \multicolumn{1}{c|}{0.05} &
  -0.04 &
  \multicolumn{1}{c|}{-0.07} &
  -0.01 &
  \multicolumn{1}{c|}{-0.01} &
  -0.01 &
  -0.04 \\ \hline
\multicolumn{1}{c|}{\multirow{3}{*}{Isolet}} &
  Balanced Accuracy &
  0.27 &
  \multicolumn{1}{c|}{0.43} &
  0.69 &
  \multicolumn{1}{c|}{0.67} &
  \textbf{0.73} &
  \multicolumn{1}{c|}{0.71} &
  0.73 &
  \multicolumn{1}{c|}{0.71} &
  0.73 &
  0.71 \\
\multicolumn{1}{c|}{} &
  F1 Score &
  0.28 &
  \multicolumn{1}{c|}{0.43} &
  0.68 &
  \multicolumn{1}{c|}{0.67} &
  0.73 &
  \multicolumn{1}{c|}{0.70} &
  0.74 &
  \multicolumn{1}{c|}{0.72} &
  \textbf{0.78} &
  0.72 \\
\multicolumn{1}{c|}{} &
  MCC &
  0.24 &
  \multicolumn{1}{c|}{0.41} &
  0.67 &
  \multicolumn{1}{c|}{0.66} &
  0.72 &
  \multicolumn{1}{c|}{0.69} &
  0.73 &
  \multicolumn{1}{c|}{0.72} &
  \textbf{0.77} &
  0.72 \\ \hline
\multicolumn{1}{c|}{\multirow{3}{*}{MADELON}} &
  Balanced Accuracy &
  0.58 &
  \multicolumn{1}{c|}{0.66} &
  0.70 &
  \multicolumn{1}{c|}{\textbf{0.81}} &
  0.78 &
  \multicolumn{1}{c|}{0.79} &
  0.78 &
  \multicolumn{1}{c|}{0.79} &
  0.78 &
  0.79 \\
\multicolumn{1}{c|}{} &
  F1 Score &
  0.58 &
  \multicolumn{1}{c|}{0.66} &
  0.70 &
  \multicolumn{1}{c|}{\textbf{0.81}} &
  0.78 &
  \multicolumn{1}{c|}{0.79} &
  0.75 &
  \multicolumn{1}{c|}{0.77} &
  0.73 &
  0.77 \\
\multicolumn{1}{c|}{} &
  MCC &
  0.16 &
  \multicolumn{1}{c|}{0.31} &
  0.40 &
  \multicolumn{1}{c|}{\textbf{0.62}} &
  0.55 &
  \multicolumn{1}{c|}{0.57} &
  0.50 &
  \multicolumn{1}{c|}{0.54} &
  0.47 &
  0.53 \\ \hline
\end{tabular}
}
\end{table}

\newpage

\section{Statistical Validation} \label{Appendix_7}
The statistical significance of results discussed in Section \ref{sec:Results} is assessed in Table \ref{tab:StatTests}. Here we report the \textit{p}-values obtained performing a $15 \times 2$cv \textit{t}-test as described in Section \ref{sec:Experiments}. Specifically, \textit{p}-values $> 0.1$ are reported in their numerical form, \textit{p}-values $\leq 0.1$ and $> 0.05$ are marked as $^{*}$, \textit{p}-values $\leq 0.05$ and $> 0.01$ are marked as $^{**}$, \textit{p}-values $\leq 0.01$ and $> 0.001$ are marked as $^{***}$ and \textit{p}-values $\leq 0.001$ are marked as $^{****}$.

\begin{table*}[ht]
\centering
\caption{Comparison between $\textrm{Inf-FS}_U$' and TFS' \textit{p}-values obtained performing a $15 \times 2$ cv paired \textit{t}-test. \textit{p}-values $> 0.1$ are reported in their numerical form, \textit{p}-values $\leq 0.1$ and $> 0.05$ are marked as $^{*}$, \textit{p}-values $\leq 0.05$ and $> 0.01$ are marked as $^{**}$, \textit{p}-values $\leq 0.01$ and $> 0.001$ are marked as $^{***}$ and \textit{p}-values $\leq 0.001$ are marked as $^{****}$. $(\vee)$ and $(\wedge)$ symbols indicate that, when the two feature selection schemes combined with the same classifier, produce statistically robust different results, TFS performs, respectively, better or worse than $\textrm{Inf-FS}_U$ according to results reported in Table \ref{tab:local_bests_results}.}
\label{tab:StatTests}
\vspace{0.1in}
\scalebox{0.6}{%
\begin{tabular}{c|ccccc|ccccc|ccccc}
\hline
            & \multicolumn{5}{c|}{\textbf{LinearSVM}} & \multicolumn{5}{c|}{\textbf{KNN}} & \multicolumn{5}{c}{\textbf{DT}}  \\ \cline{2-16} 
 &
  \textbf{10} &
  \textbf{50} &
  \textbf{100} &
  \textbf{150} &
  \textbf{200} &
  \textbf{10} &
  \textbf{50} &
  \textbf{100} &
  \textbf{150} &
  \textbf{200} &
  \textbf{10} &
  \textbf{50} &
  \textbf{100} &
  \textbf{150} &
  \textbf{200} \\ \hline
PCMAC       & *$^{(\wedge)}$ & 0.50   & 0.79  & 0.80  & 0.49  & 0.19  & 0.37 & 0.90 & 0.81 & 0.94 & *$^{(\wedge)}$   & 0.69 & 0.93 & 0.75 & 0.73 \\
RELATHE     & 0.87   & **$^{(\vee)}$     & ****$^{(\vee)}$  & ***$^{(\vee)}$   & **$^{(\vee)}$ & 0.61  & 0.17 & ****$^{(\vee)}$ & *$^{(\wedge)}$ & 0.14 & 0.90 & ***$^{(\vee)}$ & 0.58 & *$^{(\vee)}$ & 0.47 \\
COIL20      & ***$^{(\vee)}$    & ****$^{(\vee)}$   & ****$^{(\vee)}$  & ***$^{(\vee)}$   & **$^{(\vee)}$    & **$^{(\vee)}$    & **$^{(\vee)}$   & **$^{(\vee)}$   & 0.39 & 0.40 & *$^{(\vee)}$    & **$^{(\vee)}$   & ***$^{(\vee)}$  & ***$^{(\vee)}$  & **$^{(\vee)}$   \\
ORL         & 0.14   & ***$^{(\vee)}$    & 0.22  & **$^{(\vee)}$    & ***$^{(\vee)}$   & **$^{(\vee)}$    & ***$^{(\vee)}$  & 0.72 & 0.44 & 0.51 & 0.88 & 0.49 & 0.67 & **$^{(\vee)}$   & 0.93 \\
warpAR10P   & 0.23   & ***$^{(\vee)}$    & 0.38  & 0.23  & 0.20  & **$^{(\wedge)}$    & **$^{(\vee)}$   & **$^{(\vee)}$   & 0.48 & 0.69 & 0.77 & ***$^{(\vee)}$  & 0.32 & 0.16 & 0.39 \\
warpPIE10P  & 0.98   & 0.52   & 0.67  & 0.39  & 1.00  & 0.50  & 0.93 & 0.17 & 1.00 & 0.51 & 0.41 & 1.00 & 0.65 & 0.39 & 0.68 \\
Yale        & ***$^{(\vee)}$    & **$^{(\vee)}$     & 0.17  & 0.19  & 0.64  & 0.81  & **$^{(\vee)}$   & 0.28 & 0.78 & 0.10 & 0.60 & **$^{(\vee)}$   & 0.90 & 0.45 & 0.23 \\
USPS        & 0.89   & 0.38   & 0.29  & ***$^{(\vee)}$   & *$^{(\vee)}$     & 0.69  & 0.19 & ***$^{(\wedge)}$  & *$^{(\wedge)}$    & 0.26 & *$^{(\wedge)}$    & 0.74 & 0.71 & 0.62 & 0.95 \\
colon       & 0.15   & 0.42   & 0.19  & 0.37  & 0.53  & 0.92  & 0.48 & 0.63 & 1.00 & 0.18 & 0.89 & 0.35 & 0.15 & 0.59 & 0.50 \\
GLIOMA      & 0.54   & 0.63   & 0.45  & *$^{(\wedge)}$     & 0.74  & 0.68  & 0.73 & 0.25 & 0.65 & 0.91 & **$^{(\vee)}$   & 0.52 & *$^{(\vee)}$    & *$^{(\wedge)}$    & *$^{(\vee)}$    \\
lung        & ***$^{(\vee)}$    & 0.12   & ***$^{(\vee)}$   & **$^{(\vee)}$    & 0.19  & ***$^{(\vee)}$   & 0.46 & 0.61 & 0.84 & 0.58 & ***$^{(\vee)}$  & 0.11 & 0.28 & 0.47 & 0.86 \\
lung\_small & 0.72   & 0.83   & **$^{(\wedge)}$    & 0.54  & 0.32  & 0.38  & 0.77 & 0.32 & 0.82 & 0.72 & 0.73 & 0.99 & 0.53 & *$^{(\vee)}$    & 0.63 \\
lymphoma    & 0.56   & 0.26   & 0.37  & 0.74  & 0.92  & 1.00  & 0.69 & 0.41 & 0.25 & 0.86 & 0.35 & 0.58 & 0.32 & 0.45 & 0.85 \\
GISETTE     & ****$^{(\wedge)}$   & ***$^{(\wedge)}$    & ***$^{(\vee)}$   & ***$^{(\vee)}$   & **$^{(\vee)}$    & ****$^{(\vee)}$  & ****$^{(\vee)}$ & ****$^{(\vee)}$ & ****$^{(\vee)}$ & ****$^{(\wedge)}$ & ****$^{(\wedge)}$ & ***$^{(\vee)}$  & ***$^{(\wedge)}$  & **$^{(\wedge)}$   & *$^{(\vee)}$    \\
Isolet      & ****$^{(\vee)}$   & ***$^{(\vee)}$    & 0.15  & 0.15  & **$^{(\vee)}$    & ****$^{(\vee)}$  & 1.00 & 0.28 & 0.38 & 0.88 & **$^{(\vee)}$   & 0.81 & 0.61 & *$^{(\wedge)}$    & 0.39 \\
MADELON     & *$^{(\vee)}$      & 0.75   & 0.93  & 0.19  & 0.15  & 0.91  & 0.69 & 0.22 & 0.65 & 0.49 & 0.32 & 0.65 & 0.57 & 0.69 & 0.59 \\ \hline
\end{tabular}%
}
\end{table*}

\end{document}